\documentclass[11pt]{article}
\usepackage{amssymb}
\usepackage[final]{acl}
\usepackage{xurl}
\usepackage{times}
\usepackage{latexsym}
\usepackage{amsmath}
\usepackage{booktabs}
\usepackage{multirow}
\usepackage[T1]{fontenc}
\usepackage[utf8]{inputenc}
\usepackage{microtype}
\usepackage[table]{xcolor}
\usepackage{inconsolata}
\usepackage{graphicx}
\usepackage{tabularx}
\usepackage{tcolorbox}
\tcbuselibrary{skins, breakable}
\usepackage{ulem}
\usepackage{enumitem}

\definecolor{nred}{RGB}{112, 48, 160}
\definecolor{nblue}{RGB}{0, 112, 192}
\definecolor{ngreen}{RGB}{18, 141, 21}

\title{MemoNoveltyAgent: A Historical Research Memory-Aware\\Agent Workflow for Paper Novelty Assessment}
\author{
  Jiajun Hou\textsuperscript{\rm 5}\thanks{Equal contribution.} \quad
  Hexuan Deng\textsuperscript{\rm 1,3}\footnotemark[1] \quad
  Wenxiang Jiao\textsuperscript{\rm 2}
  \\
  \textbf{Xuebo Liu}\textsuperscript{\rm 1}\thanks{Corresponding author.} \quad
  \textbf{Xiaopeng Ke}\textsuperscript{\rm 1} \quad
  \textbf{Derek F. Wong}\textsuperscript{\rm 4} \quad
  \textbf{Min Zhang}\textsuperscript{\rm 1}
  \\[0.5em]
  \textsuperscript{\rm 1}Institute of Computing and Intelligence, Harbin Institute of Technology, Shenzhen, China
  \\
  \textsuperscript{\rm 2}Xiaohongshu Inc. \quad
  \textsuperscript{\rm 3}Zhongguancun Academy, Beijing, China
  \\
  \textsuperscript{\rm 4}NLP\textsuperscript{2}CT Lab, Department of Computer and Information Science, University of Macau, China
  \\
  \textsuperscript{\rm 5}The Chinese University of Hong Kong, Shenzhen, China
  \\
  \texttt{\{jiajunhou738,hxuandeng,wenxiangjiaonju,xiaopk7\}@gmail.com}
  \\
  \texttt{\{liuxuebo,zhangmin2021\}@hit.edu.cn}
  \\
  \texttt{derekfw@um.edu.mo}
}

\begin{document}
\raggedbottom
\maketitle

\begin{abstract}
To alleviate the heavy burden of paper screening, researchers increasingly rely on existing AI agents, such as AI reviewers or DeepResearch, for paper evaluation and novelty assessment. However, lacking specialized mechanisms for processing scholarly literature, their analyses often produce superficial results with noticeable deficiencies in quality. To bridge this gap, we introduce MemoNoveltyAgent, a multi-agent system designed to generate comprehensive and faithful novelty reports. Beyond retrieving concrete prior-paper evidence via RAG, our system incorporates a high-level abstract memory constructed from large-scale scholarly corpora. This memory organizes research into hierarchical trees to distill field-specific evolutionary trajectories, thereby providing a broader historical context. Furthermore, we decompose papers into discrete novelty points for fine-grained analysis and retrieval, while employing a self-validation mechanism to improve report faithfulness. Finally, to address the evaluation challenges of such open-ended generation tasks, we propose a RAG-augmented checklist evaluation method that enables reliable and evidence-grounded assessments. Extensive experiments demonstrate that MemoNoveltyAgent outperforms GPT-5 DeepResearch by 13.69\%. Code and demo are available at \url{https://github.com/SStan1/MemoNoveltyAgent}.
\end{abstract}

\section{Introduction}

Manually verifying the novelty of a paper is highly time-consuming, as it requires tracing related work and judging whether each claimed contribution is truly new. With the rapid proliferation of scientific literature, this manual overhead has increased drastically, making high-quality peer review increasingly unsustainable. Therefore, reliable automated novelty assessment has become indispensable, serving both formal peer reviewers and scholars who need to efficiently evaluate new papers.

Existing automated assistance mainly falls into two categories. General-purpose research agents, such as DeepResearch, are often repurposed for paper novelty evaluation despite being designed for general web-retrieval tasks. On the other hand, automated AI reviewers can also be employed for this assessment, though they treat novelty analysis merely as a sub-task within the broader peer-review process. However, because both approaches lack robust and sophisticated mechanisms for retrieving and processing scholarly literature, their generated analysis reports often exhibit noticeable deficiencies, particularly in terms of completeness, depth, and faithfulness, rendering the content generic, superficial, and susceptible to factual hallucinations.

These limitations arise from several primary factors. First, by processing a manuscript as an undivided whole, current workflows often obscure specific novelty points and overlook key contributions. Second, during evidence gathering, standard semantic retrieval frequently fails to retrieve earlier or implicit works that use different terminology. Third, when evaluating the gathered context, these systems rely entirely on the LLM's internal reasoning without an objective baseline. Unlike human experts who use historical literature as a reference to evaluate new submissions, LLMs lack such benchmarks, which hinders in-depth analysis and leads to generic evaluations. Finally, without explicit validation mechanisms, these systems are susceptible to hallucinations, thereby compromising the reliability of the generated analysis.

To address these limitations, we introduce MemoNoveltyAgent, a multi-agent system designed to generate comprehensive and faithful novelty reports. Specifically, MemoNoveltyAgent constructs a large-scale pre-computed abstract memory from a massive literature database. Through clustering and recursive refinement, this memory organizes ICLR research into hierarchical trees that distill the developmental histories of fine-grained subfields, thereby providing broader historical context together with expert-reviewer evaluation results. In addition, MemoNoveltyAgent builds a contextual literature database from the submission's citation network as an auxiliary reference. During evaluation, the workflow decomposes a target submission into discrete novelty points and retrieves relevant evidence from both the contextual literature database and the abstract memory. The retrieved evidence then guides report generation, followed by an independent self-validation protocol to reduce factual hallucinations and improve citation integrity. Finally, to robustly evaluate open-ended generation, we propose a checklist-based evaluation method featuring dense question coverage and a RAG-integrated verification mechanism. Extensive experiments demonstrate that MemoNoveltyAgent achieves state-of-the-art performance, outperforming the strong general-purpose baseline GPT-5 DeepResearch by 13.69\% on the main benchmark, and by 10.35\% on a broader cross-venue dataset. Our main contributions are:
\begin{itemize}
    \item We propose MemoNoveltyAgent, a multi-agent workflow for comprehensive and faithful paper novelty report generation.

    \item We build a historical research memory and combine it with citation-network-based retrieval, point-wise decomposition, and self-validation to improve report faithfulness.

    \item We introduce a RAG-integrated checklist evaluation method and validate its effectiveness and reliability through human alignment.
\end{itemize}

\section{Related Work}

\paragraph{DeepResearch.}

To overcome the inherent limitations of static parametric memory~\cite{mallen2023not,DRPruningEfficient_DJL+24}, modern LLMs such as ChatGPT~\cite{openai2024gpt4technicalreport}, Gemini~\cite{geminiteam2025geminifamilyhighlycapable}, and Kimi~\cite{moonshot2024websearch} have integrated tools like web browsing capabilities, allowing models to access external knowledge bases, thereby grounding their generation in up-to-date information~\cite{lewis2020retrieval,nakano2021webgpt,ImprovingSimultaneous_DDL+23,chen-etal-2025-sgic}.

To further address complex inquiries, ``DeepResearch'' modes have been proposed~\cite{google_gemini_deep_research,openai2025deepresearch,grok_deep_search,perplexity_deep_research}. Based on observations of their generated outputs and opinions from recent scholars~\cite{zhang2025deepresearchsurvey,shi2025deepresearchsystematicsurvey}, the core components of this paradigm generally include: first, \textit{planning}, which decomposes complex queries into sub-questions and constructs research plans~\cite{gu2024webdreamer}; second, \textit{web exploration}, where agents search for relevant information online, accompanied by continuous iteration and reflection~\cite{nakano2021webgpt,he2024webvoyager,ke-etal-2025-aquilt,rearl,DynamicSampling_RLD+25}; and finally, \textit{report generation}, which synthesizes the collected evidence into coherent responses~\cite{li2025webthinker}.

While designed for general queries and applicable to novelty assessment, their web-centric retrieval struggles with scholarly literature. Specifically, they are easily distracted by non-academic content like blogs, and their limited search breadth fails to capture the comprehensive research landscape.

\paragraph{AI Reviewer.}
Automated peer review systems aim to address the complex nature of the reviewing process~\cite{Zhuang_2025}. To capture interactive dynamics, recent agent-based frameworks simulate multi-turn dialogues~\cite{jin-etal-2024-agentreview}, treat reviewing as a long-context task~\cite{tan2024peerreviewmultiturnlongcontext}, or utilize multi-agent collaboration to enhance analytical depth~\cite{jin-etal-2024-agentreview,tian-etal-2025-agentinit,wang-etal-2026-maspo}. Furthermore, reinforcement learning aligns automated reviewers with human preferences through iterative loops~\cite{weng2025cycleresearcherimprovingautomatedresearch} or scoring optimization~\cite{zeng2025reviewrlautomatedscientificreview}, while knowledge-enhanced models provide explainable critiques~\cite{liu2024reviewrobot}. Finally, multimodal systems integrate visual encoders to interpret figures and tables for holistic assessment~\cite{Hong2025MultimodalPR}, supported by benchmarks~\cite{gao2025mmreviewmultidisciplinarymultimodalbenchmark,huang2024scimm,cocoreviewbench}.

Recent studies have begun to address scientific novelty assessment more directly. \citet{shahid2025literature} use literature retrieval and reranking to assess the novelty of scientific ideas. \citet{dasilva2025graphmind} combine structured paper representations with citation-based and semantic retrieval for interactive novelty assessment.

Despite these advances, existing systems still face certain limitations. First, directly processing the entire paper heavily increases complexity. Second, relying on LLMs to make direct judgments from raw information without references often leads to generic analyses that overlook critical details.

\section{MemoNoveltyAgent Workflow}

\begin{figure*}[t!] 
    \centering
    \includegraphics[width=\textwidth]{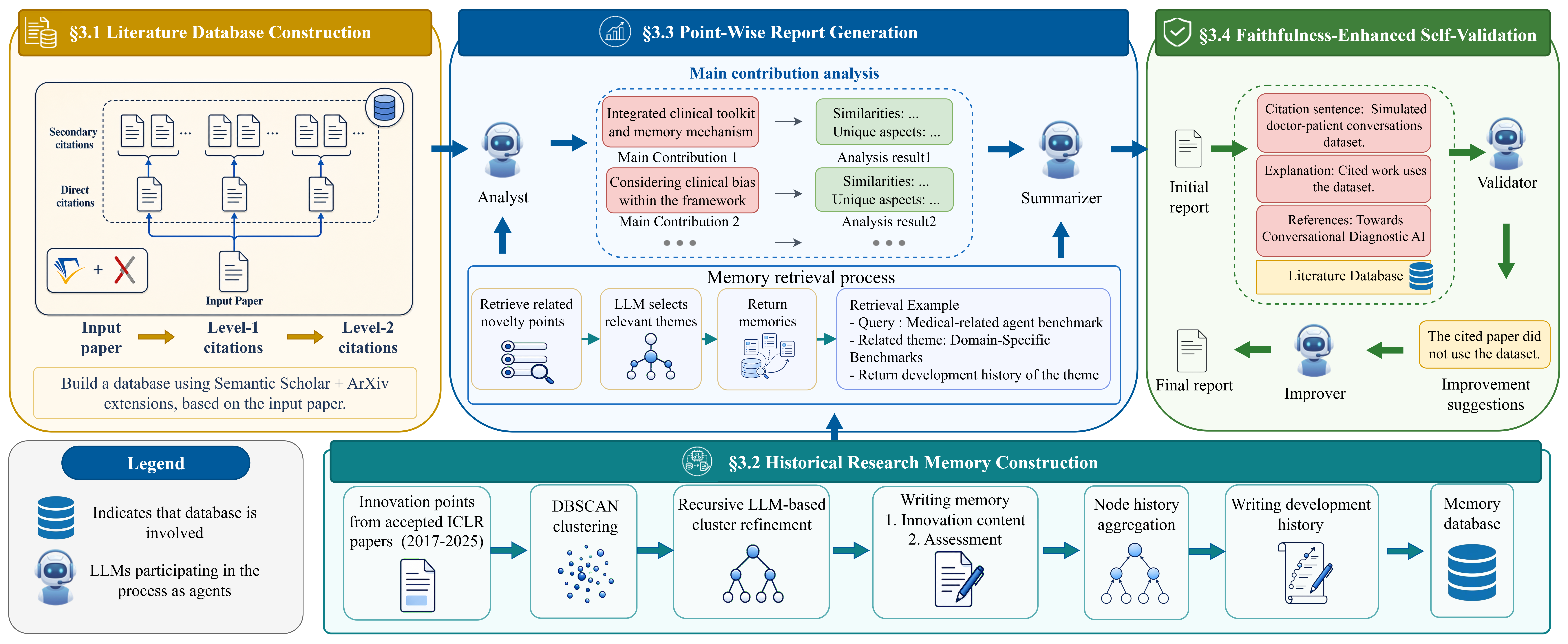} 
    \caption{The MemoNoveltyAgent workflow. The system consists of four primary modules: (\S3.1) constructing a citation-based literature database; (\S3.2) building an offline historical research memory database from ICLR accepted papers; (\S3.3) generating a point-wise report by combining local RAG and memory retrieval; and (\S3.4) performing faithfulness-enhanced self-validation to output the final report.}
    \label{fig:system_overview} 
\end{figure*}

We now detail MemoNoveltyAgent, an automated agent workflow that takes a paper title as input, downloads related content, and conducts novelty analysis. As shown in Figure~\ref{fig:system_overview}, it consists of four main stages: Literature Database Construction, Historical Research Memory Construction, Point-Wise Report Generation, and Faithfulness-Enhanced Self-Validation.

\subsection{Literature Database Construction}
\label{sec:database-construction}
Given a target paper \(P\), MemoNoveltyAgent constructs a local literature database scoped within its citation neighborhood. We begin by collecting all first-order references (i.e., the papers directly cited by \(P\)). We then expand this network to include second-order references (i.e., the papers cited by the first-order references). Because the raw second-order set can be noisy and excessively large, we prioritize papers based on structural relevance: papers cited by more first-order references are ranked higher, and more recent papers are preferred when the co-occurrence signal is tied. Through this process, each target paper is associated with a capped database of prioritized full-text PDFs. 

To facilitate precise information retrieval from this constructed database, MemoNoveltyAgent implements a hybrid Retrieval-Augmented Generation (RAG) pipeline. The detailed implementation of this RAG pipeline is provided in Appendix~\ref{app:database-rag-details}.

\subsection{Historical Research Memory Construction}

While the literature database primarily focuses on the most relevant and granular literature details, we construct an offline Historical Research Memory to provide a broader and deeper historical context. It systematically stores historical novelty points, development trajectories, and peer review comments to offer an objective reference benchmark. This module serves both as an internal knowledge anchor providing information to the LLM agents, and as a standalone, interpretable tool for researchers to directly trace the chronological development of specific fields.

\paragraph{Historical Novelty Extraction.}
We collect accepted ICLR papers from 2017 to 2025 and apply the same novelty-point extraction method used in the main workflow. This yields approximately 32K historical novelty points from more than 8K accepted papers. Each extracted point is stored alongside its source paper title, year, summary, and available OpenReview reviewer comments.

\paragraph{Coarse-to-Fine Theme Tree Construction.}
To avoid the high computational costs of relying solely on LLMs to categorize massive novelty points, we adopt a two-stage clustering strategy that balances cost and accuracy. First, we embed all novelty points and use DBSCAN to form roughly 400 coarse clusters, averaging about 80 points per cluster. Since many coarse clusters still contain multiple sub-directions, we then employ an LLM for fine-grained secondary classification. The LLM recursively refines any cluster with more than 15 items, splitting them into more granular child themes until each leaf node contains at most 15 closely related points, ultimately establishing a well-structured theme tree. A theme tree example is shown in Appendix~\ref{app:memory-details}.

\begin{figure*}[t!]
    \centering
    \includegraphics[width=\textwidth]{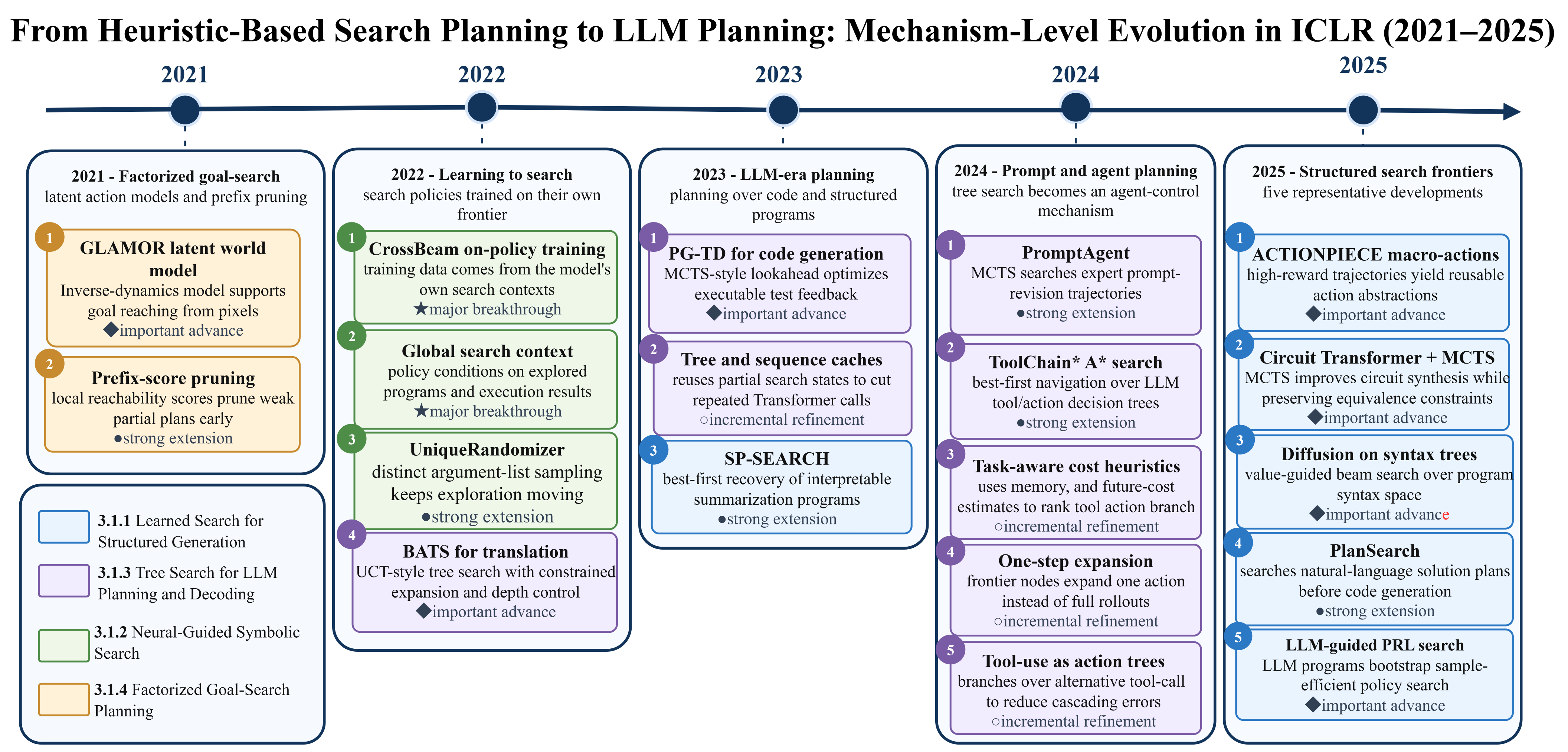}
    \caption{A concrete example of the evolutionary trajectory of a parent node.}
    \label{fig:search_based_llm_planning_timeline}
\end{figure*}

\paragraph{Writing Leaf and Aggregated Memories.}
We conceptualize each leaf node as a distinct micro-domain formed by closely related innovation points. For each micro-domain, the LLM synthesizes the innovation texts, summaries, and reviewer feedback to construct a detailed evolutionary trajectory. This record comprises two layers: (1) a specific content and novelty analysis for each individual innovation point, and (2) a chronological summary of the micro-domain's overall evolution. For parent nodes, the LLM aggregates the summaries of their child nodes to generate a broader evolutionary trajectory. Figure~\ref{fig:search_based_llm_planning_timeline} provides a concrete example of a parent node's evolutionary trajectory.

\subsection{Point-Wise Report Generation}

To ensure comprehensive evaluation and structural alignment with the Historical Research Memory, we adopt a point-by-point strategy. A dedicated agent first decomposes the manuscript into individual novelty points, which subsequent agents then analyze independently.

\paragraph{Splitting Agent.} It first extracts a set of discrete, core novelty points $\{i_1, ..., i_N\}$ from the target paper $P$. For each extracted novelty point $i_k$, the agent generates multiple targeted queries and retrieves supporting evidence from the literature database through the RAG pipeline.
\paragraph{Analyst Agent.}
Using the extracted novelty points \(i_k\) and their retrieved contexts \(\mathcal{D}_k\), the Analyst Agent conducts a rigorous comparative analysis. First, the target novelty point is embedded to retrieve the closest historical points from the Historical Research Memory, mapping them to their corresponding leaf themes. The LLM then selects the one or two most relevant themes, supplying the Analyst Agent with the specific innovation details and local analyses of all historical points within these nodes. By leveraging this macro-level organization of development trajectories, we overcome the limitations of standard semantic retrieval. This approach successfully recalls relevant papers that might otherwise be missed due to terminology variations or implicit phrasing, ensuring a comprehensive context for technical comparison.
\paragraph{Summarizer Agent.}
The Summarizer Agent synthesizes the final output and drafts the \textbf{Novelty Summary}. To distill core insights, the agent retrieves relevant context from the Historical Research Memory, specifically the novelty assessment for each innovation point, the local development history of the matched leaf theme, and the broader developmental history of its parent node. Because these development trajectories inherently incorporate historical expert peer reviews, the Summarizer Agent assimilates these critical perspectives to better simulate human expert evaluation standards during macro-summarization. Finally, it aggregates a baseline factual overview of the paper, the Point-Wise Novelty Analysis, and the Novelty Summary to construct the complete \textbf{Initial Report}.

\subsection{Faithfulness-Enhanced Self-Validation}

To ensure report credibility and mitigate hallucinations, MemoNoveltyAgent incorporates a self-validation module that verifies and corrects the generated content against the source materials.

\paragraph{Validator and Improver Agents.} 
To prevent source distortion caused by the lack of independent verification, the \textbf{Validator Agent} systematically constructs \textbf{citation pairs}, linking report statements directly to their full-text sources. It scrutinizes these pairs for strict faithfulness and generates actionable feedback for any identified misinterpretations. Guided by this feedback, the \textbf{Improver Agent} performs directional corrections to rectify hallucinations. Following these targeted edits, it conducts a comprehensive polishing pass to enhance linguistic fluency, terminological consistency, and logical flow, ultimately guaranteeing the rigorous faithfulness of the final analysis.
\paragraph{Output Report.}
The final output report \(R\) follows a structured, three-section format. First, the Paper Content Summary provides a concise factual overview of \(P\), capturing its core objectives and key technical approaches. Second, the Point-wise Novelty Analysis conducts a comparative analysis against retrieved prior work for each extracted novelty point, identifying technical similarities and highlighting the genuine innovations. Finally, the Novelty Summary synthesizes these findings into a dialectical assessment that carefully distinguishes genuine breakthroughs from incremental modifications. Specific cases are showcased in Appendix~\ref{report-example}.

\subsection{Checklist-based Evaluation}

Evaluating open-ended novelty reports is challenging: manual assessment is unscalable, and standard metrics like BLEU~\cite{papineni-etal-2002-bleu} and ROUGE~\cite{lin-2004-rouge} fail to capture semantic depth and factual accuracy. Even recent LLM-as-a-judge approaches struggle to provide reliable evaluations for such nuanced tasks~\cite{li2026evaluatingscoringbiasllmasajudge,wang-etal-2024-large-language-models-fair}. To address this, we develop an automated, \textbf{RAG-enhanced checklist-based evaluation method}. By decomposing holistic assessment into specific, verifiable criteria (Yes/No questions), this approach forces the model to make judgments based on atomic, concrete facts, thereby minimizing subjectivity and enhancing reliability~\cite{cook2024ticking}. 

\paragraph{Checklist Construction.}
Drawing inspiration from CheckEval~\cite{lee-etal-2025-checkeval}, we construct an evaluation checklist spanning five key dimensions: Fluency assesses linguistic quality and clarity of expression; Faithfulness evaluates factual alignment with source materials to prevent hallucinations; Completeness measures whether the report omits any essential information that should be considered; Effectiveness confirms strict adherence to the predefined template and focus on core analysis; and Depth evaluates intellectual rigor and the inclusion of specific technical details. We construct this checklist, comprising a total of 118 questions, through LLM generation followed by human verification. The detailed construction process is provided in Appendix~\ref{app:checklist-construct}. We then compute, for each model, the proportion of checklist questions answered with yes and rescale this proportion to a 0--10 range as the model's final score.
\paragraph{RAG Integration.} 
For faithfulness and other criteria requiring external evidence, we connect the evaluator to the database constructed in \S\ref{sec:database-construction}, enabling the LLM to retrieve relevant paper content during verification. Detailed prompts and implementation specifics are provided in Appendices~\ref{app:database-rag-details} and~\ref{app:checklist-construct}. Furthermore, by integrating a RAG system, this evaluation framework can be extended to a wider range of scenarios, providing a viable paradigm for assessing open-ended text generation in the future.

\begin{table*}[t]
\centering
\small
\setlength{\tabcolsep}{8pt}
\renewcommand{\arraystretch}{0.92}

\begin{tabular}{lcccccc}
\toprule
\multirow{2}{*}{\textbf{Models}} & \multicolumn{6}{c}{\textbf{Score (0-10) $\uparrow$}} \\
\cmidrule(lr){2-7}
& \textbf{Completeness} & \textbf{Depth} & \textbf{Effectiveness} & \textbf{Faithfulness} & \textbf{Fluency} & \textbf{Overall} \\
\midrule
\multicolumn{7}{l}{\textit{Web Research}} \\
\midrule
Kimi-2 & 7.57 & 7.01 & 9.22 & 4.85 & 9.77 & 7.68 \\
GPT-5 Thinking & 7.92 & 7.21 & 9.67 & 4.90 & 9.87 & 7.91 \\
Gemini-2.5-Flash DeepResearch & 8.18 & 8.03 & 9.58 & 4.00 & 8.61 & 7.68 \\
GPT-5 DeepResearch & 8.06 & 7.72 & 9.45 & 6.06 & 9.59 & 8.18 \\
\midrule
\multicolumn{7}{l}{\textit{AI Reviewer}} \\
\midrule
DeepReview & 8.39 & 8.10 & 9.51 & 7.56 & 9.85 & 8.68 \\
AgentReviewer & 8.55 & 8.07 & 9.28 & 6.46 & \cellcolor{green!15}\textbf{9.91} & 8.45 \\ \midrule
MemoNoveltyAgent & \cellcolor{green!15}\textbf{9.20} & \cellcolor{green!15}\textbf{8.80} & \cellcolor{green!15}\textbf{9.93} & \cellcolor{green!15}\textbf{8.76} & 9.83 & \cellcolor{green!15}\textbf{9.30} \\
\bottomrule
\end{tabular}
\caption{Main evaluation results on the 50-paper ICLR dataset (0--10 scale). Best results are highlighted.}
\label{tab:main-50-positive}
\end{table*}

\begin{table}[t]
  \footnotesize
  \centering
  \setlength{\tabcolsep}{1pt}
  \renewcommand{\arraystretch}{0.92}
  \begin{tabular}{lcccc}
    \toprule
    \multirow{2}{*}{\textbf{Models}} &
    \multicolumn{2}{c}{\textbf{Proxy ($n{-}1$)}} &
    \multicolumn{2}{c}{\textbf{Proxy ($n$)}} \\
    \cmidrule(lr){2-3}\cmidrule(lr){4-5}
    & \textbf{MSE}$\downarrow$ & \textbf{MAE}$\downarrow$ & \textbf{MSE}$\downarrow$ & \textbf{MAE}$\downarrow$ \\
    \midrule
    Gemini-2.5-Flash-Nothinking & \textbf{0.28} & \textbf{0.44} & \textbf{0.19} & \textbf{0.37} \\
    Gemini-2.5-Flash & 0.37 & 0.49 & 0.26 & 0.41 \\
    GPT-4o-Mini & 0.72 & 0.79 & 0.50 & 0.66 \\
    Claude-3-Haiku-20240307 & 0.93 & 0.77 & 0.64 & 0.64 \\
    DeepSeek-V3-250324 & 1.11 & 0.99 & 0.77 & 0.82 \\
    O4-Mini-2025-04-16 & 1.46 & 1.06 & 1.01 & 0.88 \\
    \bottomrule
  \end{tabular}
  \caption{LLM Consistency result on a 32-paper subset.}
  \label{tab:proxy_review_compare}
\end{table}

\section{Experiments}

To evaluate the performance of our method, we design our experiments to answer the following research questions: \textbf{RQ1}: How does MemoNoveltyAgent perform compared to existing automated tools? \textbf{RQ2}: How reliable and effective is the proposed automated evaluation method? \textbf{RQ3}: What is the individual contribution of each core component, and how well does the framework generalize across different scenarios?

\subsection{Experimental Setup}

\paragraph{Dataset Construction.}
We constructed an evaluation dataset based on ICLR 2025 submissions from OpenReview. To ensure diversity and representativeness, we select 50 target papers via stratified sampling based on a weighted combination of reviewer overall scores and originality scores. These target papers are evenly distributed across four specific score ranges (0-3, 3-5, 5-7, and 7-10) to ensure a balanced evaluation. Furthermore, the selection maximizes the coverage of research fields, encompassing papers from 20 distinct research topics. For each target paper, we constructed a literature database comprising 200 PDF files.

\paragraph{Baseline Models.}
We compare MemoNoveltyAgent against two categories of baselines: \textit{Web Research} and \textit{AI Reviewer}. The web research systems include GPT-5 Thinking, Kimi-2~\cite{moonshot2024websearch}, GPT-5 DeepResearch~\cite{openai2025deepresearch}, and Gemini-2.5-Flash DeepResearch~\cite{google_gemini_deep_research}. These systems retrieve online information to generate novelty reports, with some of them supporting more advanced deep research capabilities such as multi-step web exploration. For all web research systems, we conduct testing through the services provided on their official websites. The AI reviewer systems include AgentReviewer~\cite{jin-etal-2024-agentreview} and DeepReview~\cite{zhu2025deepreviewimprovingllmbasedpaper}, which use agent-based paper reviewing frameworks and proprietary academic databases to generate novelty reports.

\begin{table*}[t]
  \centering
  \small
  \setlength{\tabcolsep}{6pt}
  \renewcommand{\arraystretch}{0.92}
  \begin{tabular}{lcccccccccccc}
    \toprule
    \multirow{3}{*}{\textbf{Models}} & \multicolumn{12}{c}{\textbf{Score (1--5) $\uparrow$}} \\
    \cmidrule(lr){2-13}
    & \multicolumn{2}{c}{\textbf{Completeness}} & \multicolumn{2}{c}{\textbf{Depth}} & \multicolumn{2}{c}{\textbf{Effectiveness}} & \multicolumn{2}{c}{\textbf{Faithfulness}} & \multicolumn{2}{c}{\textbf{Fluency}} & \multicolumn{2}{c}{\textbf{Overall}} \\
    \cmidrule(lr){2-3} \cmidrule(lr){4-5} \cmidrule(lr){6-7} \cmidrule(lr){8-9} \cmidrule(lr){10-11} \cmidrule(lr){12-13}
    & H & AI & H & AI & H & AI & H & AI & H & AI & H & AI \\
    \midrule
    DeepReview & 3.875 & 4.17 & 3.50 & 4.16 & 4.75 & 4.78 & 3.75 & 4.04 & \cellcolor{green!15}\textbf{4.875} & \cellcolor{green!15}\textbf{4.91} & 4.15 & 4.41 \\
    GPT-5 DeepResearch & 3.50 & 4.15 & 3.25 & 3.99 & 4.50 & 4.67 & 3.00 & 3.15 & 4.50 & 4.78 & 3.75 & 4.15 \\
    MemoNoveltyAgent & \cellcolor{green!15}\textbf{4.25} & \cellcolor{green!15}\textbf{4.53} & \cellcolor{green!15}\textbf{4.50} & \cellcolor{green!15}\textbf{4.34} & \cellcolor{green!15}\textbf{4.88} & \cellcolor{green!15}\textbf{4.89} & \cellcolor{green!15}\textbf{4.25} & \cellcolor{green!15}\textbf{4.32} & 4.50 & 4.87 & \cellcolor{green!15}\textbf{4.48} & \cellcolor{green!15}\textbf{4.59} \\
    \bottomrule
  \end{tabular}
  \caption{Human (H) and AI evaluation result on a 32-paper subset. AI scores are scaled to the 1--5 human rubric.}
  \label{tab:human_ai_eval}
\end{table*}
\subsection{Main Results}

Table~\ref{tab:main-50-positive} presents the main evaluation results across the five defined dimensions on our primary testbed. MemoNoveltyAgent achieves the highest overall score of 9.30, outperforming the strongest baseline, DeepReview, by a margin of 0.62 points. 

\paragraph{Factual Accuracy and Reliability.}
MemoNoveltyAgent achieves the highest Faithfulness score of 8.76, demonstrating that our generated results are highly reliable with minimal hallucinations. In contrast, web-based baselines generally exhibit lower faithfulness as they are susceptible to data volatility and unverified claims. 

\paragraph{Analytical Depth and Critical Review Perspectives.}
MemoNoveltyAgent achieves the highest Depth score of 8.80. This indicates that our system provides richer technical details and objective evaluations. Unlike baselines that often generate generic summaries from isolated text, our framework retrieves fine-grained technical features. Furthermore, by incorporating expert review perspectives, the system generates more professional and critical technical assessments.

\paragraph{Breadth of Literature Coverage.}
MemoNoveltyAgent secures a leading Completeness score of 9.20. This evaluation result underscores that our generated reports suffer from fewer omissions and achieve broader literature coverage. Standard baselines frequently miss crucial prior works due to evolving terminologies across different eras. Our framework alleviates this bottleneck through a dual-source retrieval mechanism, which couples the local literature database with historical technological trajectories in our memory module, maximizing information recall.

\paragraph{Structural and Linguistic Proficiency.}
Regarding structural compliance and readability, the evaluated models generally maintain high performance standards, with most models scoring above 9.20 in both dimensions. This indicates that current large language models are inherently capable of satisfying baseline linguistic fluency and format adherence. Consequently, the critical performance differentiation among automated reviewing systems is driven primarily by factual correctness and analytical thoroughness rather than surface presentation. 
\begin{table*}[t]
  \centering
  \small
  \setlength{\tabcolsep}{6pt}
  \renewcommand{\arraystretch}{0.92}
  \begin{tabular*}{\textwidth}{@{\extracolsep{\fill}}lcccccc@{}}
    \toprule
    \multirow{2}{*}{\textbf{Models}} & \multicolumn{6}{c}{\textbf{Score (0-10) $\uparrow$}} \\
    \cmidrule(lr){2-7}
    & \textbf{Completeness} & \textbf{Depth} & \textbf{Effectiveness} & \textbf{Faithfulness} & \textbf{Fluency} & \textbf{Overall} \\
    \midrule
    MemoNoveltyAgent & \cellcolor{green!15}\textbf{9.20} & \cellcolor{green!15}\textbf{8.80} & \cellcolor{green!15}\textbf{9.93} & \cellcolor{green!15}\textbf{8.76} & 9.83 & \cellcolor{green!15}\textbf{9.30} \\
    \quad w/o Memory & 8.76 & 8.49 & 9.79 & 8.43 & 9.66 & 9.03 \\
    \quad Direct Generation & 8.35 & 7.44 & 8.29 & 7.78 & \cellcolor{green!15}\textbf{9.86} & 8.34 \\
    \bottomrule
  \end{tabular*}
  \caption{Ablation results on the main 50-paper ICLR dataset. ``w/o Memory'' only removes the memory module. ``Direct Generation'' removes memory, point-wise analysis, and self-validation.}
  \label{tab:ablation-50}
\end{table*}
\begin{table*}[t]
  \centering
  \small
  \setlength{\tabcolsep}{6pt}
  \renewcommand{\arraystretch}{1.05}
  \begin{tabular}{lcccc}
    \toprule
    \textbf{Method} & \textbf{Completeness} & \textbf{Depth} &
    \textbf{Reviewer Helpfulness} & \textbf{Faithfulness} \\
    \midrule
    Kimi-2 Online & 3.06 & 3.14 & 3.02 & 4.74 \\
    GPT-5 Thinking & 3.42 & 3.42 & 3.44 & 4.88 \\
    Gemini-2.5-Flash DeepResearch & 3.40 & 3.62 & 3.40 & 4.98 \\
    GPT-5 DeepResearch & 3.66 & 3.82 & 3.68 & 4.98 \\
    AgentReviewer & 4.00 & 4.04 & 3.98 &
    \cellcolor{green!15}\textbf{5.00} \\
    DeepReview & 4.00 & 4.22 & 4.02 & 4.96 \\
    MemoNoveltyAgent &
    \cellcolor{green!15}\textbf{4.47} &
    \cellcolor{green!15}\textbf{4.85} &
    \cellcolor{green!15}\textbf{4.60} & 4.98 \\
    \bottomrule
  \end{tabular}
  \caption{Reviewer-reference evaluation against an approximate ground truth constructed from novelty-related comments in the original ICLR 2025 OpenReview reviews. Scores are reported on a 1--5 scale, and the best result in each dimension is highlighted.}
  \label{tab:expert_reference}
\end{table*}

\subsection{Evaluation Method Validity}

We aim to validate the effectiveness of our proposed evaluation method. Due to the absence of an absolute gold-standard answer, we employ two distinct validation approaches: human alignment and LLM consistency. Inspired by recent automated benchmarking methodologies, the LLM consistency approach utilizes multiple language models as evaluators, taking their average consensus score as a proxy gold standard. If our evaluation method demonstrates high alignment with human judgments and exhibits minimal deviation from the LLM consensus, it is considered robust and reliable.

\paragraph{LLM Consistency.}
To rigorously evaluate the consistency of our automated evaluation method, we conducted a study on a representative subset of 32 papers. Specifically, we collected the novelty reports generated by four baselines---GPT-5 Online, GPT-5 DeepResearch, MemoNoveltyAgent, and DeepReview---to serve as the evaluation targets. To mitigate potential systematic bias stemming from a single model architecture, we deliberately selected evaluator models from four distinct vendors, as illustrated in Table~\ref{tab:proxy_review_compare}. To establish proxy ground truths (GT) for quantifying scoring consistency, we employed two distinct aggregation strategies: (1) the \(n-1\) strategy, where the proxy GT for a specific evaluator model is defined as the average score assigned by the remaining models; and (2) the \(n\) strategy, where the proxy GT is calculated by averaging the scores from all participating models. We then computed the Mean Absolute Error (MAE) and Mean Squared Error (MSE) to quantify the mathematical alignment between each model's predicted score and the collective proxy GT.

\paragraph{High Consistency Across Models.}
Table~\ref{tab:proxy_review_compare} presents the comparative results of various evaluator models. Notably, Gemini-2.5-Flash-Nothinking emerges as the top-performing model, achieving a remarkably low MSE of \textbf{0.28} and an MAE of \textbf{0.44} under the \(n-1\) strategy. This superior alignment indicates that the model closely adheres to the collective consensus. Driven by this exceptional stability, we adopt it as the primary evaluator for all subsequent formal assessments, which further substantiates that our overall evaluation method yields stable and reliable outcomes.

\paragraph{Human Evaluation Alignment.}
To verify if our automated proxy metrics anchor well with professional human standards, we sampled 8 papers from each score interval to build a 32-paper subset for manual grading. As detailed in Table~\ref{tab:human_ai_eval}, human experts and our automated checklist evaluation method exhibit remarkably tight alignment. From a pairwise comparison perspective, except for language fluency which results in a statistical tie, the preference rankings between the automated evaluator and human experts are completely consistent across all other core dimensions. Compared to the automated evaluator, human reviewers observe a wider margin in analytical depth and demonstrate higher acceptance for MemoNoveltyAgent. This strongly validates the effectiveness of our automated evaluation method and its high alignment with human cognition.

\paragraph{Reviewer-Reference Evaluation.}
\label{app:expert-reference}

Beyond comparing automated scores with human ratings, we directly evaluate report content against an external reference constructed from real reviewer comments. This experiment covers all 50 papers in the main ICLR 2025 dataset. For each paper, GPT-5.4 reads all original OpenReview reviews and extracts novelty-related comments as structured points, including contributions recognized as novel and aspects considered insufficiently novel or incremental. We treat these structured reviewer points as an approximate ground truth for the novelty issues that a useful report should address.

We then use GPT-5.4 to compare each complete generated report against these extracted reviewer comments. The presentation order of reports from different systems is randomized to reduce position bias. Each report is evaluated on a 1--5 scale along four dimensions: \textit{Completeness}, measuring coverage of reviewer-identified novelty comments; \textit{Depth}, measuring whether the report discusses the same fine-grained technical mechanisms; \textit{Reviewer Helpfulness}, measuring how effectively the report reduces the need for manual literature retrieval and comparison; and \textit{Faithfulness}, measuring whether the report contains explicit factual conflicts with the reviewer comments.

\begin{table*}[t]
  \centering
  \small
  \setlength{\tabcolsep}{6pt}
  \renewcommand{\arraystretch}{0.92}
  \begin{tabular*}{\textwidth}{@{\extracolsep{\fill}}lcccccc@{}}
    \toprule
    \multirow{2}{*}{\textbf{Models}} & \multicolumn{6}{c}{\textbf{Score (0--10) $\uparrow$}} \\
    \cmidrule(lr){2-7}
    & \textbf{Completeness} & \textbf{Depth} & \textbf{Effectiveness} & \textbf{Faithfulness} & \textbf{Fluency} & \textbf{Overall} \\
    \midrule
    GPT-5.4 Online & 6.30 & 3.87 & 9.49 & 5.18 & \cellcolor{green!15}\textbf{9.73} & 6.91 \\
    GPT-5.4 DeepResearch & 7.64 & 7.42 & 9.66 & 6.91 & 9.40 & 8.21 \\
    DeepReview & 8.42 & 7.72 & 9.79 & 7.30 & 9.54 & 8.55 \\
    MemoNoveltyAgent & \cellcolor{green!15}\textbf{8.82} & \cellcolor{green!15}\textbf{8.50} & \cellcolor{green!15}\textbf{9.94} & \cellcolor{green!15}\textbf{8.57} & 9.49 & \cellcolor{green!15}\textbf{9.06} \\
    \bottomrule
  \end{tabular*}
  \caption{Evaluation result on the cross-venue dataset, including 50 papers from ACL, CVPR, ACM MM, TPAMI, and JMLR.}
  \label{tab:cross_venue}
\end{table*}

As shown in Table~\ref{tab:expert_reference}, MemoNoveltyAgent ranks first in Completeness, Depth, and Reviewer Helpfulness, achieving scores of 4.47, 4.85, and 4.60, respectively. These results show that its reports more fully cover reviewer-identified novelty issues, provide more technically detailed comparisons, and offer greater practical support for reviewers. Faithfulness scores are uniformly high across systems and differ only slightly, partly because original reviewer comments are typically concise and provide limited information from which explicit factual conflicts can be identified. Overall, this experiment compares the generated reports against novelty-related assessments provided by human reviewers as an approximate ground truth. The resulting rankings are consistent with those of the preceding evaluations, thereby complementing the existing validation analyses and making the overall evaluation framework more comprehensive.

\subsection{Ablation Study and Component Analysis}

To assess the distinct performance contribution of each component, we conducted incremental ablation experiments on our main dataset. Table~\ref{tab:ablation-50} details the quantitative variations.

\paragraph{Impact of Base Framework Architecture.}
Compared to ``Direct Generation'', adopting the base agentic framework (w/o Memory) increases the overall score from 8.34 to 9.03, reflecting the structural advantages of our workflow. This base framework improves performance primarily through two core mechanisms. First, the notable improvement in Faithfulness (+0.65) benefits from the closed-loop self-validation mechanism, which effectively filters hallucinations. Second, the increases in Depth (+1.05) and Completeness (+0.41) are driven by the point-wise analysis paradigm. By decomposing the overall report generation process into independent, fine-grained novelty points, the system retrieves richer context to provide specific technical details and ensure broader coverage of relevant information.

\paragraph{Impact of Historical Research Memory.}
Integrating the historical research memory pushes the overall rating from 9.03 to its peak of 9.30. The empirical benefits of this module are reflected in two dimensions. For literature Completeness (+0.44), the memory acts as a vital supplementary source alongside the literature database, establishing a robust dual-source retrieval mechanism that ensures broader information recall. For analytical Depth (+0.31), the incorporation of historical peer-review feedback equips the framework with an expert reference benchmark. Rather than relying solely on raw paper texts, this professional baseline calibrates the language model's cognitive scale, thereby significantly enhancing its independent, dialectical reasoning and critical thinking capabilities.
\subsection{Cross-Venue Generalization Analysis}

To evaluate the performance of MemoNoveltyAgent across different research domains and writing styles, we establish a cross-venue dataset.

\paragraph{Cross-Venue Dataset Setup.}
We construct a cross-venue dataset consisting of 50 papers sampled evenly  from five venues: ACL, CVPR, ACM MM, TPAMI, and JMLR. It covers natural language processing, computer vision, multimodal learning, and statistical machine learning, spanning both conference and journal publication formats. 

\paragraph{Performance Stability.}
Table~\ref{tab:cross_venue} shows the cross-venue evaluation results. MemoNoveltyAgent achieves the highest overall score of 9.06, outperforming DeepReview by 0.51 points, and maintains a consistent lead across completeness (8.82), depth (8.50) and faithfulness (8.57). While absolute scores show a slight overall decline compared to the native ICLR environment---primarily because our historical memory is constructed using previously accepted ICLR papers, which do not strictly align with the distributions of papers from these other venues, resulting in cross-venue domain shifts---the framework maintains reliable reporting performance without specialized fine-tuning. 

\section{Conclusion}

In this work, we present MemoNoveltyAgent, a multi-agent workflow for analyzing the novelty of academic papers and generating comprehensive, faithful reports. At its core, the system introduces a historical research memory assisted by a point-wise analysis and self-validation architecture. To support a reliable evaluation of this open task, we also propose a checklist-based evaluation method. We hope MemoNoveltyAgent can serve as a reliable tool for high-quality novelty analysis, reducing the cost of paper screening and helping researchers quickly identify truly original papers.

\section*{Limitations}
We acknowledge a few limitations in our current approach. First, our proposed agent framework relies heavily on open-access academic papers, which is why our current evaluation and application are primarily focused on the Computer Science (CS) domain. In several other disciplines, such as biological sciences, a substantial portion of academic literature is behind paywalls and lacks public retrieval methods. This makes it exceedingly difficult to acquire a sufficient volume of full-text papers required to construct the historical research memory and the local literature database. To extend our framework to these restricted domains, future work will need to explore alternative technical routes (e.g., partnering with academic publishers or developing lightweight verification strategies based solely on abstracts and metadata). Second, for every target input, our framework dynamically constructs a local literature database, which necessitates downloading a large volume of PDF files in real time. This process imposes a significant demand on network resources. Consequently, on servers or environments with poor network connectivity, the PDF acquisition phase can consume a considerable amount of time, thereby bottlenecking the overall efficiency. Finally, the execution time for processing a single paper is relatively long. The complex multi-agent workflow requires each agent to occupy a substantial amount of processing time for extensive retrieval, reading, and reasoning. However, this limitation is primarily evident in sequential, single-instance runs; when processing large batches of papers, the overall time cost can be effectively mitigated through parallel execution and distributed deployment.

\section*{Acknowledgments}
This work is supported in part by the Zhongguancun Academy (Grant No.s C20250203), the Science and Technology Development Fund of Macau SAR (Grant Nos. FDCT/0007/2024/AKP, EF2024-00185-FST), the UM and UMDF (Grant Nos. MYRG-GRG2024-00165-FST-UMDF, MYRG-GRG2025-00236-FST), the Tencent AI Lab Rhino-Bird Research Program (Grant No. EF2023-00151-FST), the Stanley Ho Medical Development Foundation (Grant No. SHMDF-AI/2026/001), and the National Natural Science Foundation of China (Grant No. 62266013).

\section*{Ethics Statement}
Our work adheres to the ACL Ethics Policy and uses only publicly available scholarly resources and datasets for reproducibility. Since the data come from open-access academic resources, we do not collect private data from individuals. For the human annotation involved in this work, participants were informed of the purpose of the study and how their annotations would be used before participation. LLMs may exhibit racial and gender biases, so we strongly recommend users assess potential biases before applying the system in specific contexts. Additionally, due to the difficulty of controlling LLM outputs, users should be cautious about hallucinations and unreliable generations.

\bibliography{custom}

\newpage
\appendix
\section{Appendix}

\label{sec:appendix}

\subsection{Case Study}
\label{app:case-study}

Although quantitative evaluation captures broad differences across systems, illustrating how our framework avoids generic statements and whether the generated reports approach the phrasing of human experts is often difficult to demonstrate directly through quantitative metrics alone. Therefore, we provide specific qualitative examples in Table~\ref{tab:case_study} for supplementary explanation.

\paragraph{Reducing generic comparisons.}
The point-wise analysis process helps reduce generic, high-level paraphrasing and encourages the model to recover concrete technical details from the paper and related work. As shown in the first type of case study in Table~\ref{tab:case_study}, GPT-5 DeepResearch describe a related medical AI system merely as one that ``simulated patient-doctor conversations,'' whereas our report identifies the more specific OSCE-style consultation setting and connects it to history-taking, diagnostic reasoning, and communication skill evaluation. Similarly, for mathematically oriented papers, our framework does not merely state that a paper uses ``topology and category theory,'' but successfully identifies the first derived functor, quiver-representation resolutions, and the resulting path-count formulation. These examples fully demonstrate that decomposed novelty analysis can encompass more technical details, thereby effectively avoiding generic statements.

\paragraph{Memory-enabled expert-like judgment.}
Human reviews provided by historical memory serve as a reference, enabling the model's analysis to more closely align with a human expert's perspective, particularly in accurately defining the scope and level of the proposed contributions. As demonstrated in the second type of case study, while the baseline model without memory often provides generic praise or broad endorsements, our memory-augmented model can explicitly point out the exact nature of a contribution. For instance, in the clinical evaluation case, the report positions the work as an ``integrative and domain-specialized'' advancement rather than a ``transformative methodological breakthrough.'' Furthermore, in the concept bottleneck model (CBM) case, the report notes that the individual building blocks are not unprecedented, but their systematic adaptation to the CBM framework is what makes it valuable. These calibrated expressions help precisely define the scope and level of the innovations, making the phrasing much closer to the professional standard of real human reviewers.

\definecolor{detailgreen}{RGB}{34, 139, 34}
\definecolor{labelgray}{RGB}{120, 120, 120}

\begin{table*}[t]
\centering
\small
\begin{tabularx}{\textwidth}{l X}
\toprule
\multicolumn{2}{p{\dimexpr\textwidth-2\tabcolsep}}{\textbf{Case Type 1: Generic Statements vs. Fine-Grained Technical Details}} \\
\midrule

\textit{Context} & Evaluating the specific technical alignment and similarities with AMIE. \\
GPT-5 DeepResearch & AMIE also simulated patient-doctor conversations to improve diagnosis. \hfill \textcolor{labelgray}{\textit{Overly generic}} \\
\textbf{MemoNoveltyAgent} & Resembling the proposed approach, AMIE utilized \textcolor{detailgreen}{\textbf{OSCE-style simulated consultations}} for training and evaluation to \textcolor{detailgreen}{\textbf{assess history-taking, diagnostic reasoning, and communication skills.}} \\

\midrule
\textit{Context} & Evaluating whether the report explains the mathematical mechanism behind a proposed emergence metric. \\
GPT-5 DeepResearch & The paper provides a formal measure of emergence using topology and category theory. \hfill \textcolor{labelgray}{\textit{High-level restatement}} \\
\textbf{MemoNoveltyAgent} & The paper identifies emergence with the \textcolor{detailgreen}{\textbf{first derived functor}} of an observation functor, computes it through \textcolor{detailgreen}{\textbf{projective/injective resolutions in quiver representations}}, and reduces the result to a \textcolor{detailgreen}{\textbf{path-count formula}} over affected network edges. \\

\midrule
\midrule
\multicolumn{2}{p{\dimexpr\textwidth-2\tabcolsep}}{\textbf{Case Type 2: Monolithic Paraphrasing vs. Calibrated Expert Judgment}} \\
\midrule

\textit{Context} & Evaluating the nature and scope of a paper's core contributions. \\
Baseline \textit{(w/o Memory)} & The manuscript presents a well-motivated, non-obvious integration that meaningfully advances multimodal, sequential clinical evaluation. \hfill \textcolor{labelgray}{\textit{Generic praise}} \\
\textbf{MemoNoveltyAgent} & The component techniques are adapted from existing agent work, so the contribution is \textcolor{detailgreen}{\textbf{integrative and domain-specialized}} rather than a \textcolor{detailgreen}{\textbf{transformative methodological breakthrough.}} \\

\midrule
\textit{Context} & Evaluating a deployment-oriented LLM routing and abstention framework. \\
Baseline \textit{(w/o Memory)} & The system-level design and Pareto analysis constitute a substantial and distinctive contribution. \hfill \textcolor{labelgray}{\textit{Mostly affirmative}} \\
\textbf{MemoNoveltyAgent} & The framework is \textcolor{detailgreen}{\textbf{deployment-ready and operationally valuable}}, but its individual components mainly \textcolor{detailgreen}{\textbf{refine rather than revolutionize}} selective evaluation, calibration, and model-routing literature. \\

\midrule
\textit{Context} & Evaluating the originality of an editable concept bottleneck model framework. \\
Baseline \textit{(w/o Memory)} & The three-level taxonomy and influence-function formulation are treated as genuine methodological contributions. \hfill \textcolor{labelgray}{\textit{Less calibrated}} \\
\textbf{MemoNoveltyAgent} & The individual building blocks are not unprecedented, but the \textcolor{detailgreen}{\textbf{systematic adaptation to concept bottleneck models is valuable}} because it organizes editing, correction, and concept-level intervention into a coherent CBM-specific framework. \\

\midrule
\textit{Context} & Evaluating whether a new sequence model architecture is a breakthrough or a targeted synthesis. \\
Baseline \textit{(w/o Memory)} & The architecture is described as a substantial design-level contribution. \hfill \textcolor{labelgray}{\textit{Broad endorsement}} \\
\textbf{MemoNoveltyAgent} & The contribution is better understood as a \textcolor{detailgreen}{\textbf{targeted architectural synthesis}} rather than a wholesale departure, since it builds on prior Mamba, Liquid, S3, and SPACETIME-style ideas while adapting them to the paper's specific modeling objective. \\

\midrule
\textit{Context} & Evaluating the contribution of a modified Transformer attention operator. \\
Baseline \textit{(w/o Memory)} & The report presents the method as a cohesive algorithmic, theoretical, and empirical contribution. \hfill \textcolor{labelgray}{\textit{Less discriminating}} \\
\textbf{MemoNoveltyAgent} & The core novelty is a \textcolor{detailgreen}{\textbf{concrete operator-level change}} in the attention mechanism, while the accompanying stability techniques are mostly \textcolor{detailgreen}{\textbf{supporting engineering choices}} rather than independent conceptual breakthroughs. \\

\bottomrule
\end{tabularx}
\caption{Case studies illustrating report quality. Case Type 1 shows that point-wise analysis helps recover concrete technical details rather than generic comparisons. Case Type 2 shows that historical research memory helps calibrate novelty judgments by distinguishing genuine contributions from domain-specific integrations, targeted refinements, or supporting engineering choices.}
\label{tab:case_study}
\end{table*}

\subsection{Historical Research Memory Implementation and Example}
\label{app:memory-details}
\begin{figure*}[t]
    \centering
    \includegraphics[width=\textwidth]{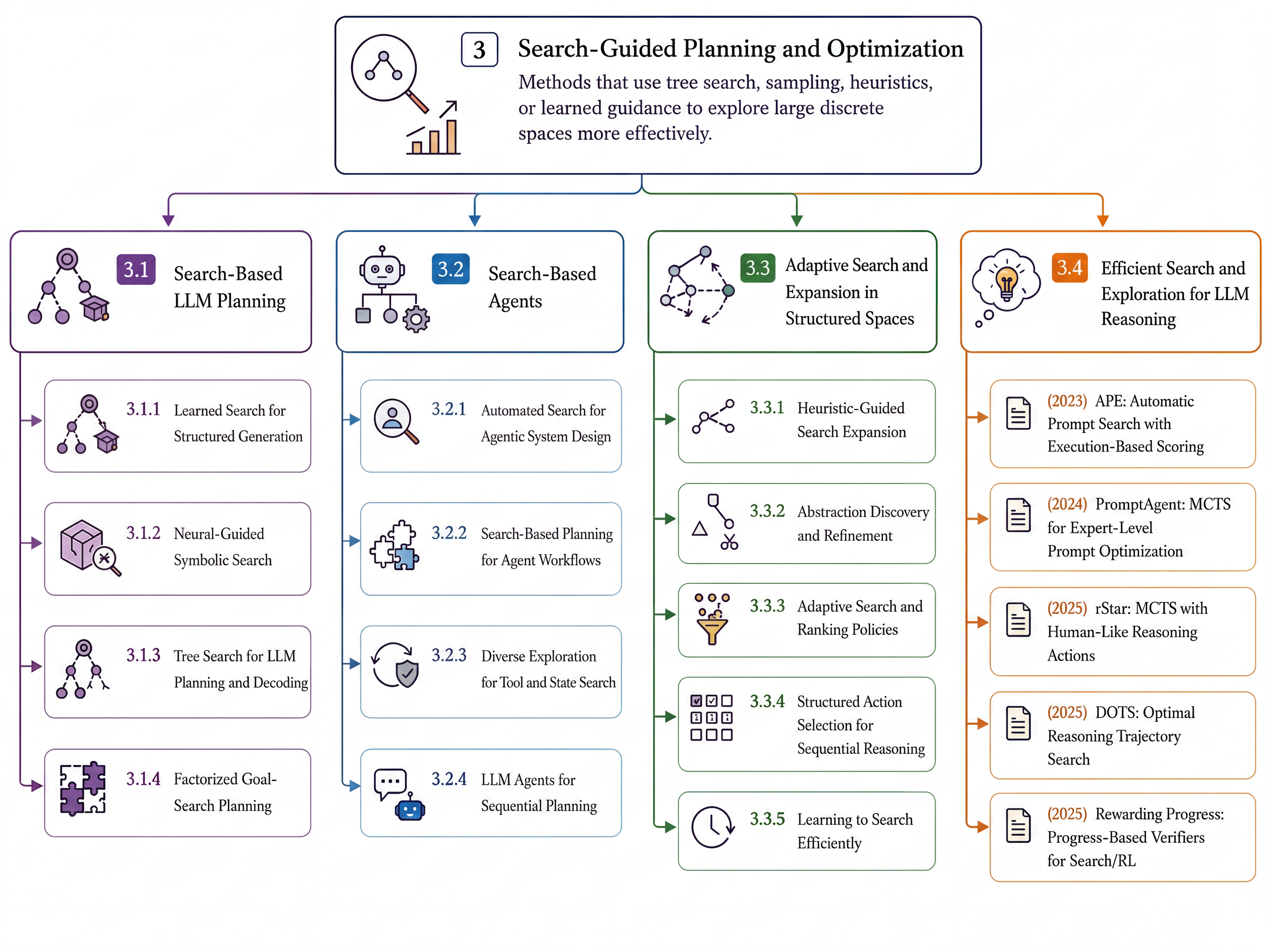}
    \caption{An example of theme tree from the historical research memory: a clustered development tree for the search-guided planning and optimization theme. Branches are recursively refined only when the corresponding theme exceeds the refinement threshold; therefore, some branches are further divided into subthemes, while others directly list papers.}
    \label{fig:theme-tree-example}
\end{figure*}

This appendix mainly presents a concrete example of the historical research memory, specifically illustrating the generated theme tree. Rather than repeating the construction pipeline already described in the main text, Figure~\ref{fig:theme-tree-example} provides one produced memory branch for the theme \textit{Search-Guided Planning and Optimization}. The figure should be read as a data-driven development tree over extracted novelty points: the top node is a broad theme, intermediate nodes are subthemes produced by clustering and LLM-based refinement, and terminal leaves are either fine-grained subthemes or individual papers attached to a sufficiently small theme.

In this example, node 3 splits into several research paths. Branch 3.1, for instance, is further divided into 3.1.1--3.1.4 because its clustered innovation points still contain multiple distinguishable directions, such as learned search for structured generation, neural-guided symbolic search, tree search for LLM planning and decoding, and factorized goal-search planning. By contrast, branch 3.4 directly lists papers because the branch itself contains fewer than 15 innovation records under the refinement rule, so the construction pipeline does not split it into additional child themes. This mixed structure is expected: the memory tree is not a manually balanced taxonomy, but a hierarchy whose depth reflects the local density and diversity of historically extracted innovation points.

To facilitate public access and exploration, the entire historical memory database is deployed as an interactive website, which is provided alongside our open-source code. Users can leverage this web interface to intuitively navigate the global memory structure. For example, users can utilize keyword searches to locate a specific node within the database and examine its adjacent nodes. Additionally, they can view the specific development history of the current node itself, or explore the broader evolutionary trajectory of its parent aggregated nodes.

\subsection{Cross-Venue Breakdown}
\label{app:cross-venue-breakdown}
Table~\ref{tab:cross_venue_breakdown} provides the detailed scores broken down by each conference or journal for the cross-venue generalization experiment. Across these diverse venues (ACL, ACM MM, CVPR, JMLR, and TPAMI), MemoNoveltyAgent demonstrates solid overall results and remains the best system in every venue. The margin is not identical across venues, but the pattern is stable: MemoNoveltyAgent improves completeness, depth, and faithfulness in most venue-level slices, while fluency remains comparable to strong baselines.

Because the historical memory is constructed based on ICLR papers, directly transferring it to these other venues causes a slight decrease in the absolute scores. Nevertheless, MemoNoveltyAgent still consistently outperforms other baselines, maintaining a clear advantage.

\begin{table*}[t]
\centering
\footnotesize
\setlength{\tabcolsep}{3pt}
\renewcommand{\arraystretch}{1.05}
\begin{tabular}{llcccc}
\toprule
\textbf{Venue} & \textbf{Dimension} & \textbf{DeepReview} & \textbf{MemoNoveltyAgent} & \textbf{GPT-5.4 DeepResearch} & \textbf{GPT-5.4 Online} \\
\midrule
\multirow{6}{*}{ACL}
& Completeness & 8.61 & 9.07 & 7.84 & 6.00 \\
& Depth & 7.07 & 8.15 & 7.74 & 3.48 \\
& Effectiveness & 9.77 & 9.94 & 9.82 & 9.59 \\
& Faithfulness & 6.48 & 7.93 & 6.79 & 5.79 \\
& Fluency & 9.86 & 9.00 & 9.50 & 9.79 \\
\rowcolor{green!10}
& Overall & 8.36 & 8.82 & 8.34 & 6.93 \\
\midrule
\multirow{6}{*}{ACM MM}
& Completeness & 8.65 & 9.42 & 8.10 & 6.23 \\
& Depth & 7.96 & 9.04 & 7.89 & 3.41 \\
& Effectiveness & 9.82 & 10.00 & 9.47 & 9.47 \\
& Faithfulness & 7.59 & 8.48 & 6.83 & 5.31 \\
& Fluency & 9.21 & 9.86 & 9.71 & 9.71 \\
\rowcolor{green!10}
& Overall & 8.65 & 9.36 & 8.40 & 6.83 \\
\midrule
\multirow{6}{*}{CVPR}
& Completeness & 8.58 & 8.74 & 7.52 & 6.55 \\
& Depth & 8.04 & 8.70 & 7.59 & 3.96 \\
& Effectiveness & 9.82 & 10.00 & 9.65 & 9.71 \\
& Faithfulness & 7.24 & 9.10 & 6.72 & 5.69 \\
& Fluency & 9.50 & 9.57 & 9.43 & 9.71 \\
\rowcolor{green!10}
& Overall & 8.64 & 9.22 & 8.18 & 7.12 \\
\midrule
\multirow{6}{*}{JMLR}
& Completeness & 7.84 & 8.16 & 7.42 & 6.58 \\
& Depth & 7.44 & 7.93 & 7.07 & 4.63 \\
& Effectiveness & 9.71 & 10.00 & 9.77 & 9.47 \\
& Faithfulness & 7.41 & 9.00 & 7.35 & 3.86 \\
& Fluency & 9.29 & 9.71 & 9.00 & 9.86 \\
\rowcolor{green!10}
& Overall & 8.34 & 8.96 & 8.12 & 6.88 \\
\midrule
\multirow{6}{*}{TPAMI}
& Completeness & 8.42 & 8.71 & 7.32 & 6.16 \\
& Depth & 8.07 & 8.67 & 6.78 & 3.85 \\
& Effectiveness & 9.82 & 9.77 & 9.59 & 9.24 \\
& Faithfulness & 7.79 & 8.31 & 6.86 & 5.24 \\
& Fluency & 9.86 & 9.29 & 9.36 & 9.57 \\
\rowcolor{green!10}
& Overall & 8.79 & 8.95 & 7.98 & 6.81 \\
\bottomrule
\end{tabular}
\caption{Detailed venue-level breakdown for cross-venue evaluation (0--10 scale).}
\label{tab:cross_venue_breakdown}
\end{table*}

\subsection{Database and Retrieval Details}
\label{app:database-rag-details}
To construct the local literature database, we expand the target paper's first-order references to include second-order citations. To manage the noise and scale of this expanded set, we prioritize papers based on structural relevance (i.e., co-occurrence among first-order references) and recency. This targeted selection is based on a reasonable assumption: recent papers with high local exposure in a specific research neighborhood are more likely to capture its core directions and to overlap with the target paper in novelty. They therefore constitute the most important potential sources of novelty conflict. Furthermore, we incorporate the full-text content of these prioritized papers rather than relying solely on abstracts. This allows the agent to access intricate technical details, thereby facilitating a significantly deeper and more granular analysis of potential novelty conflicts.

Following PDF parsing, we explicitly remove reference lists to minimize noise. The cleaned text is then divided into chunks of 512 tokens. Each chunk is encoded into a dense vector representation using the \textit{maidalun1020/bce-embedding-base\_v1} model and indexed in a vector database. During the retrieval phase, the LLM generates targeted search queries for each novelty point, dynamically tailored to specific task requirements and contextual nuances. To ensure comprehensive coverage, exactly 6 queries are generated per novelty point: 4 queries composed of relevant keywords and 2 queries formulated as specific contextual sentences. Inspired by RAGFlow\footnote{\url{https://github.com/infiniflow/ragflow}}, we adopt a two-stage hybrid retrieval approach. The first stage conducts initial recall through two parallel pathways: a sparse retrieval path based on BM25 lexical matching to capture exact terms (e.g., method names), and a dense retrieval path based on vector similarity to capture semantic relevance. The scores from both pathways are normalized and fused. In the second stage, the top-ranked candidates from the hybrid recall are forwarded to the Qwen3-Reranker-4B cross-encoder model~\cite{zhang2025qwen3embeddingadvancingtext} for fine-grained relevance estimation. Finally, the reranker selects the top 7 most relevant chunks for each query, which are subsequently provided to the Analyst Agent—and the checklist evaluator—as highly relevant, citation-grounded evidence.

\subsection{Citation Accuracy}
For Citation Accuracy (CA), we follow the core idea of DeepResearchBench~\cite{du2025deepresearchbenchcomprehensivebenchmark}: citation-bearing claims are extracted from the report, mapped to the cited full-text papers, and judged according to whether the cited source actually supports the claim. Under this protocol, self-validation improves CA from 89.90 to 93.72, while GPT-5 DeepResearch obtains 87.60. This indicates that the Validator and Improver agents do not merely polish the report, but materially reduce unsupported or misattributed citation claims.

\begin{table*}
\centering
\small
\setlength{\tabcolsep}{12pt} 
\renewcommand{\arraystretch}{1.05}
\begin{tabular}{lccc}
\toprule
\textbf{Dimension} & \textbf{Better (Ours)} & \textbf{Worse (Ours)} & \textbf{Tie} \\
\midrule
Completeness & 100.0\% & 0.0\% & 0.0\% \\
Depth & 100.0\% & 0.0\% & 0.0\% \\
Effectiveness & 100.0\% & 0.0\% & 0.0\% \\
Faithfulness & 100.0\% & 0.0\% & 0.0\% \\
Fluency & 100.0\% & 0.0\% & 0.0\% \\
\midrule
\rowcolor{green!10}
\textbf{Overall Win Rate} & \textbf{100.0\%} & \textbf{0.0\%} & \textbf{0.0\%} \\
\bottomrule
\end{tabular}
\caption{Pairwise comparison results between our generated reports and OpenNovelty reports on 10 randomly selected ICLR 2026 submissions. ``Better'' indicates the percentage of cases where our report was preferred.}
\label{tab:opennovelty_comparison}
\end{table*}

\subsection{System Efficiency and Cost Analysis}
\label{app:efficiency-cost}

Whenever a new target paper is evaluated, the system needs to crawl data to construct a tailored reference dataset. The exact speed of this process depends heavily on the network conditions of the environment running the agent system. To evaluate the system's efficiency, we selected 10 papers to measure the time cost. The primary time consumption in our pipeline comes from the chunking and embedding of the reference literature, as well as the RAG retrieval process. The remaining time consists of LLM API calls, which can be easily and significantly accelerated through parallelization.

On average, for each paper, the RAG retrieval process takes approximately 30.5 seconds, while the chunking and embedding process takes about 701.3 seconds. For comparison, the GPT series' deep research baseline in our study takes an average of about 1100 seconds. Furthermore, it occasionally suffers from stalling issues, requiring up to 1 hour and 15 minutes to resolve in such edge cases. In contrast, baselines relying solely on simple web search are much faster, taking only about 25 seconds.

In terms of API cost, our work is highly economical. Because we utilize a smaller, cost-effective model (GPT-5-mini) and perform comparisons based on RAG-retrieved content (i.e., comparing relevant retrieved text chunks rather than feeding the entire full texts into the context window), the average cost is only \$0.075 per paper according to OpenAI's API pricing. In contrast, the GPT series' deep research baseline is expected to incur a significantly higher cost, although it is difficult to calculate its exact token consumption directly. 

Overall, our approach strategically saves on expensive resources, such as API calls and GPU compute, by relying primarily on much cheaper resources like network bandwidth and CPU processing. While this trade-off may result in a longer processing time per paper, the overall financial cost remains relatively low. Furthermore, since these network and CPU-bound processes are highly parallelizable, our system can be easily scaled up for efficient, large-scale execution.

Additionally, in our open-source version, users can easily adjust the size of the reference database for each paper (the default is set to 200 related papers). If a user reduces this number to 50, the database construction time drops to approximately 180 seconds, which can effectively improve the overall efficiency of the system based on user needs.

\clearpage

\subsection{Comparison with OpenNovelty}
\label{app:comparison-opennovelty}

OpenNovelty\cite{zhang2026opennoveltyllmpoweredagenticverifiable} is highly relevant to our research, as it also conducts in-depth investigations into the novelty of academic papers. However, it cannot be functionally reproduced at this time because its open-source repository relies on a private API that has not yet been publicly released. To conduct a comparison, we randomly selected 10 reports generated for ICLR 2026 submissions from their publicly available examples and compared their quality with our generated reports.

To minimize the bias caused by the structural differences between our reports and theirs, we adopted a pairwise comparison method following previous work\cite{zheng2023judgingllmasajudgemtbenchchatbot}. Furthermore, we utilized the same scoring criteria provided to our human judges as the rubric to guide the LLM's assessment. As shown in Table~\ref{tab:opennovelty_comparison}, our reports outperformed OpenNovelty across all evaluated dimensions.

However, for the sake of fairness, we must point out that the core objective of OpenNovelty is fundamentally different from ours. OpenNovelty focuses on making a binary (yes or no) judgment on whether a proposed innovation conflicts with prior work, providing clear, traceable evidence to assist human reviewers, without involving further AI-driven analysis or reasoning. Although its results are presented in the form of a report, the comprehensive quality of the report text is not its absolute core focus. In contrast, our evaluation criteria are primarily based on the overall quality of the generated reports, which explains why our results are completely superior to theirs under this specific evaluation framework.

\subsection{Output Report Example}
\label{report-example}
The following presents an output example of the AgentClinic report. Due to space limitations, this appendix only provides the complete Section 1, a representative novelty point from Section 2, and selected excerpts from Section 3. Omitted portions are explicitly marked.

\subsection{Human Evaluation Instructions}
To validate the automated evaluation method, we asked experienced Master's students specializing in artificial intelligence to score generated reports on a 1--5 scale. The rubric below translates the manual scoring guide used by annotators and keeps it in the same full-width framed format as the prompt examples.

\subsection{Evaluation Checklist and Core Prompts}
\label{app:checklist-construct}
The following figures present the core prompts used in the evaluation and report-generation pipelines. For readability, repetitive instructions are omitted, while the task definitions, input fields, constraints, and output formats are retained.

\paragraph{Evaluation Checklist Items.}
The implemented checklist contains 118 yes/no questions: 14 for Fluency, 17 for Effectiveness, 31 for Completeness, 29 for Faithfulness, and 27 for Depth.

\normalsize
\begin{figure*}
\centering
\begin{minipage}{0.97\textwidth}
\begin{tcolorbox}[enhanced,colback=gray!10,colframe=black,coltitle=white,fonttitle=\bfseries,title={Output Report Example: Paper Content Summary},boxrule=1pt,arc=2pt,left=6pt,right=6pt,top=6pt,bottom=6pt,fontupper=\footnotesize]
\textbf{1. Paper Content Summary} \\
This paper introduces AgentClinic, a publicly released, multimodal agent benchmark designed to evaluate large language (and vision) models in simulated clinical environments that require interactive, dialogue-driven, sequential decision making rather than static question-answering. It aims to address the gap that conventional medical benchmarks (e.g., MedQA/USMLE) present all information up-front and do not capture iterative history-taking, test-ordering, multimodal image interpretation, tool use, bias effects, multilingual needs, specialist domains, or patient-centered outcomes. To do so, the authors implement four LLM-based agents (doctor, patient, measurement/instrument reader, and moderator) and populate cases from USMLE/MedQA, deidentified EHRs (MIMIC-IV), NEJM case challenges, and specialist datasets (MedMCQA). The benchmark supports nine specialties, seven languages, and 23 cognitive/implicit bias scenarios injected via agent role prompts. Doctor agents can use an "agent toolbox" of capabilities including chain-of-thought prompting (zero/one-shot/reflection), adaptive retrieval-augmented generation (RAG) from textbooks or web sources, an experience-persistent notebook for experiential learning, and multimodal image requests.

Evaluation compares multiple LLM backbones (Claude-3.5, GPT-4/4o/3.5, Mixtral-8x7B, Llama variants) across AgentClinic-MedQA, AgentClinic-MIMIC-IV, and multimodal NEJM cases using metrics beyond diagnostic accuracy. These include patient-agent confidence, compliance, consultation willingness, and clinician dialogue realism/empathy ratings. Key findings reported are that sequential, interactive diagnosis is substantially harder than static MedQA (diagnostic accuracy can fall dramatically and MedQA performance is only weakly predictive of AgentClinic performance), Claude-3.5 often attains the highest accuracy across settings, models differ markedly in their ability to benefit from tools (the notebook yields large gains for some models such as Llama-3), GPT-4 shows relative robustness to instructed biases while models like Mixtral degrade more, and simulated biases can reduce patient-agent trust and compliance.

Definitions: Objective Structured Clinical Examination (OSCE) is a dialogue-driven clinical skills exam template used to structure cases; Retrieval-Augmented Generation (RAG) denotes retrieving external documents to ground model generations; Sparse Mixture-of-Experts (SMoE) is an architecture that routes tokens to a subset of expert blocks (used by Mixtral); USMLE is the United States Medical Licensing Exam; NEJM refers to the New England Journal of Medicine.
\end{tcolorbox}
\end{minipage}
\caption{Paper content summary excerpted from the AgentClinic report.}
\end{figure*}

\begin{figure*}
\centering
\begin{minipage}{0.97\textwidth}
\begin{tcolorbox}[enhanced,colback=gray!10,colframe=black,coltitle=white,fonttitle=\bfseries,title={Output Report Example: Point-wise Novelty Analysis},boxrule=1pt,arc=2pt,left=6pt,right=6pt,top=6pt,bottom=6pt,fontupper=\footnotesize]
\textbf{2.1. Novelty Point 1: (Classification: Dataset/Benchmark)} \\
\textbf{a) Claimed Novelty.} Explanation: AgentClinic is presented as an open-source dataset/benchmark suite that simulates longitudinal, dialogue-driven clinical encounters in an OSCE-style framework, extending static medical QA into interactive, multimodal clinical workflows. Concretely, AgentClinic composes case JSONs (patient/doctor/measurement/moderator agents and persistent state) by drawing on USMLE/MedQA items, NEJM case challenges, and de-identified EHR records (MIMIC-IV). The benchmark explicitly supports multimodal inputs (clinical text plus medical images), interactive actions (ordering/requesting tests, invoking image reads, performing measurements), persistent documentation (notes that carry across the encounter), and patient-centric follow-ups (confidence, compliance, consultation ratings). It is organized into variants (AgentClinic-MedQA, -MIMIC-IV, -NEJM, -Spec, -Lang) covering nine specialties and seven languages, and the paper reports large-scale experiments across seven LLM backbones, example model accuracies (e.g., Claude-3.5 \textasciitilde{}62.1\% +/-3.3 on AgentClinic-MedQA), multimodal NEJM evaluations, specialty/language breakdowns, and clinician-rated realism/empathy scores -- demonstrating both the resource and its empirical utility.

\textbf{b) Similarities.} - The cited AMIE work presents an approach that develops conversational diagnostic AI using simulated OSCE-style interactions: it builds a simulated clinician/patient role-play and self-play simulated-dialogue pipeline to train and evaluate diagnostic dialogue systems, and it evaluates AMIE via randomized OSCE-style text consultations with specialist physician and patient-actor assessments. [1] - Use of simulated patient-physician conversations and OSCE-like data-generation procedures resembles earlier datasets and work that create simulated clinical dialogues for training/evaluation. [2] - Building on existing medical QA benchmarks (USMLE/MedQA/MedMCQA) as source material follows established practice in medical-LM evaluation. [3] \textit{...}

\textbf{c) Unique Differences.} - AgentClinic combines multimodal clinical inputs (text + medical images) with an OSCE-style, sequential, tool-enabled clinical workflow in a single open benchmark -- prior medical OSCE-style evaluations were predominantly text/chat-based and prior multimodal medical LLM work rarely packaged into an interactive clinical-encounter benchmark. [1] [4] - The benchmark's explicit case JSON schema that separates patient/doctor/measurement/moderator agents, supports persistent notes and measurements, and exposes explicit test-ordering and image-read actions (i.e., structured tool-like clinical operations) is not matched by earlier medical dialogue datasets or by general agent benchmarks that are not specialized for clinical semantics. [2] [6] \textit{...}

\textbf{d) Details of Unique Differences.} As a dataset/benchmark, AgentClinic fills a concrete gap: prior OSCE-style medical evaluation work demonstrated that LLM-based systems can be assessed on multi-turn diagnostic dialogue with clinician and patient ratings, but those studies were largely text-only and focused on diagnostic accuracy and communication metrics within simulated dialogue systems; they did not simultaneously provide a standardized, multimodal, tool-enabled benchmark that (1) encodes clinical actions (test orders, image reads, persistent EHR-like notes) as first-class operations, (2) sources cases from both standardized exam items (USMLE/MedQA) and high-complexity clinical narratives (NEJM) plus real EHR data (MIMIC-IV), and (3) includes multilingual and multi-specialty partitions with patient-centered follow-ups. The practical consequence is that AgentClinic enables evaluation of capabilities that matter for deployed clinical agents -- sequential decision-making under information-gathering constraints, multimodal image+text interpretation in the loop, persistent documentation and state, and patient-facing metrics -- all in one reproducible suite. This combination meaningfully extends prior diagnostic-dialogue and multimodal-medical-LM work by converting isolated capabilities into an integrated, clinically-grounded benchmark that can measure whether models coordinate information acquisition, tool use, and communication across an encounter -- a capability set that earlier medical benchmarks and general agent suites either tested separately or did not operationalize for clinical semantics. [1] [4] [5]

\textit{Additional novelty points in the source report are omitted only for space.}
\end{tcolorbox}
\end{minipage}
\caption{Point-wise novelty analysis excerpted from the AgentClinic report.}
\end{figure*}

\begin{figure*}[t]
\centering
\begin{minipage}{0.97\textwidth}
\begin{tcolorbox}[enhanced,colback=gray!10,colframe=black,coltitle=white,fonttitle=\bfseries,title={Output Report Example: Novelty Summary},boxrule=1pt,arc=2pt,left=6pt,right=6pt,top=6pt,bottom=6pt,fontupper=\footnotesize]
\textbf{3. Novelty Summary} \\
\textbf{Overall characterization.} - This work assembles an integrated, reproducible clinical agent-evaluation stack that moves beyond isolated static medical QA by combining OSCE-style multi-turn encounters, multimodal inputs (images + text), explicit tool-like clinical operations (test ordering, image reads, persistent documentation), agent-role specialization, and configurable bias injections -- and uses that stack to run controlled, factorial experiments measuring diagnostic accuracy, tool effects, bias provenance, and patient-centred outcomes. The core contribution is an applied, domain-specialized evaluation and experimental platform that operationalizes several agent-evaluation desiderata previously explored separately into a single clinical workflow.

\textbf{Genuine innovations (excerpted).} - What is new and how it is realized: The benchmark packages clinical-case JSON templates that explicitly encode patient/doctor/measurement/moderator agents, persistent encounter state (notes, measurements), requestable diagnostic instruments, and patient follow-ups, while sourcing cases across exam-style items, high-complexity narrative cases, and de-identified EHR rows and offering multilingual and specialty partitions. It therefore converts static medical QA items into reproducible, sequential clinical tasks with multimodal stimuli and downstream patient-impact signals. \textit{...} - What is new and how it is realized: The system treats measurement instruments as first-class agents (request -> instrument output -> clinician interpretation), enforces role-specific information exposure and constrained inquiry budgets, provides runtime tool access (adaptive retrieval, CoT/reflection variants), and implements a persistent notebook that carries experience across cases -- all evaluated and ablated empirically. The measurement-as-agent and persistent cross-case notebook combined with clinical inquiry budgets instantiate a realistic diagnostic test loop that prior general multi-agent frameworks did not operationalize for clinical semantics at scale. \textit{...}

\textbf{Aspects lacking substantial originality (excerpted).} - Use of OSCE-style simulated dialogues and self-play simulated patient/clinician role-play -- Category: Standard Practice. Improvement suggestion: incorporate prospectively collected clinician--patient recorded interactions or clinician-in-the-loop adjudication to further validate and diversify dialogue dynamics. \textit{...} - Sourcing questions/cases from exam items (USMLE/MedQA) and NEJM case challenges -- Category: Standard Practice. Improvement suggestion: add protocols and leakage analyses to guarantee no contamination with model-pretraining corpora and prioritize clinician-authored novel vignettes for higher ecological novelty. \textit{...}

\textbf{Final One-line Summary.} \textbf{3 - Good / Significant:} This paper delivers a significant, well-executed advance by packaging multimodal, tool-enabled OSCE-style clinical evaluation, a domain-specialized multi-agent simulation (including measurement-as-agent and persistent notebook), and a clinically focused bias-injection and empirical study suite -- providing an important resource and actionable empirical findings for clinical AI assessment -- while most component techniques are adapted from existing agent and bias-evaluation work so the contribution is integrative and domain-specialized rather than a transformative methodological breakthrough.
\end{tcolorbox}
\end{minipage}
\caption{Novelty summary excerpted from the AgentClinic report.}
\end{figure*}

\begin{figure*}[p]
\centering
\begin{minipage}{0.97\textwidth}
\begin{tcolorbox}[enhanced,colback=gray!10,colframe=black,coltitle=white,fonttitle=\bfseries,title={Human Evaluation Rubric: Fluency, Effectiveness, and Completeness},boxrule=1pt,arc=2pt,left=6pt,right=6pt,top=6pt,bottom=6pt,fontupper=\footnotesize]
\setlength{\parindent}{0pt}\setlength{\parskip}{0.15em}
\textbf{Fluency.} Core criteria: grammatical correctness, concise expression, readability, and absence of repetition. \textbf{5 - Excellent:} The report is extremely fluent; all sentences are grammatically correct; capitalization is proper; the content is concise; there is no cross-section repetition or redundant restatement of technical details; terminology is consistent; and the reading experience is excellent. \textbf{4 - Good:} The report is generally fluent and grammatical, with at most isolated redundant expressions or slightly verbose sentences; the overall structure is compact and terminology is mostly consistent. \textbf{3 - Fair:} The report is readable but contains visible verbosity or repetition, and may contain minor grammar/spelling errors or overly complex sentences. \textbf{2 - Poor:} The report is difficult to read because of multiple grammar or formatting issues, repeated content, improper wording, or confused terminology. \textbf{1 - Very Poor:} The report is almost unreadable, logically confused, filler-heavy, and not aligned with academic-report writing standards.

\textbf{Effectiveness.} Core criteria: strict three-section template compliance, logical coherence, and no irrelevant content. \textbf{5 - Excellent:} The report strictly follows the three-section structure of content summary, comparative analysis, and novelty summary, without extra headings or sections; Section 2 contains the required a/b/c/d items with coherent logic; Section 3 is fully grounded in earlier analysis and includes the required 1--4 score and one-sentence summary; the report stays strictly on the assigned paper. \textbf{4 - Good:} The report follows the three-section structure and satisfies the main formatting requirements, with at most minor subsection-title deviations or a slightly abrupt Section 3 summary. \textbf{3 - Fair:} The overall structure is roughly correct but has visible format violations, such as a missing required subsection or a broken logic chain. \textbf{2 - Poor:} The structure is confused, the three main parts are not clearly separated, or the report contains substantial irrelevant domain history/background; the required score or summary is missing. \textbf{1 - Very Poor:} The report ignores the template or becomes a generic essay unrelated to the assigned paper-analysis task.

\textbf{Completeness.} Core criteria: coverage of the main paper's core elements, all novelty points, and related-paper comparisons. \textbf{5 - Excellent:} There are no meaningful omissions; the summary includes the target paper's objective, methods, and uncommon abbreviations; the comparative analysis identifies all similar innovation points mentioned in related literature and fully analyzes similarities and unique differences for every main point; the reported differences are genuinely unique. \textbf{4 - Good:} The report covers most core content and main novelty points, but some secondary comparison details may be insufficient or some similarities may be missed. \textbf{3 - Fair:} The report has clear omissions, such as missing a key method, omitting an important related paper, or claiming a difference that has an obvious similar prior work. \textbf{2 - Poor:} The report misses key parts, ignores major novelty points, omits many similar papers, or gives non-unique "unique" differences. \textbf{1 - Very Poor:} The report provides very little information, omits most retrieved evidence, and gives insufficient analysis.
\end{tcolorbox}
\end{minipage}
\caption{Human evaluation rubric, part 1.}
\label{fig:human-rubric-part1}
\end{figure*}

\begin{figure*}[p]
\centering
\begin{minipage}{0.97\textwidth}
\begin{tcolorbox}[enhanced,colback=gray!10,colframe=black,coltitle=white,fonttitle=\bfseries,title={Human Evaluation Rubric: Faithfulness and Depth},boxrule=1pt,arc=2pt,left=6pt,right=6pt,top=6pt,bottom=6pt,fontupper=\footnotesize]
\setlength{\parindent}{0pt}\setlength{\parskip}{0.15em}
\textbf{Faithfulness.} Core criteria: grounding in retrieved text, no hallucination, correct attribution, and no exaggeration. \textbf{5 - Excellent:} The content is entirely based on the provided target-paper and related-paper text; all similarities and differences have clear textual evidence; the report does not exaggerate achievements; all factual claims are grounded; and there is no misunderstanding or contradiction. \textbf{4 - Good:} Most content is accurate, with at most slight ambiguity in a secondary detail and no core factual error or serious hallucination. \textbf{3 - Fair:} The report contains questionable statements, partial misunderstanding, distorted meaning, or claims that are only inferred rather than directly supported. \textbf{2 - Poor:} The report contains clear hallucinations or errors, such as fabricated data/comparisons or attributing related-paper content to the target paper. \textbf{1 - Very Poor:} The report contradicts or is unrelated to the retrieved text and contains unsupported speculation or fabrication.

\textbf{Depth.} Core criteria: concrete technical terminology, quantitative or specific evidence, and independent critical thinking. \textbf{5 - Excellent:} The analysis is highly detailed, uses specific names rather than vague descriptions, cites concrete results to support claims, and shows independent thinking by identifying limitations or uncertainty and giving evidence-based novelty judgments. \textbf{4 - Good:} The report includes major technical terms and some concrete comparisons and can identify mechanism-level differences, but may lack quantitative support; the analysis is mostly independent. \textbf{3 - Fair:} The report remains at the level of functional description, says that something is better or different without explaining why in technical detail, and mostly restates the original content. \textbf{2 - Poor:} The report is superficial, uses vague promotional adjectives without technical support, largely copies author claims, and lacks independent judgment. \textbf{1 - Very Poor:} The report is empty or vague, misses the core technical contribution, cannot reason independently, and may contradict earlier analysis.
\end{tcolorbox}
\end{minipage}
\caption{Human evaluation rubric, part 2.}
\label{fig:human-rubric-part2}
\end{figure*}
\clearpage

\begin{figure*}[p]
\centering
\begin{minipage}{0.97\textwidth}
\begin{tcolorbox}[enhanced,colback=gray!10,colframe=black,coltitle=white,fonttitle=\bfseries,title={Evaluation Checklist: Fluency (14 questions)},boxrule=1pt,arc=2pt,left=6pt,right=6pt,top=6pt,bottom=6pt,fontupper=\footnotesize]
\setlength{\parindent}{0pt}\setlength{\parskip}{0pt}
\textbf{1.} Is the report free from formatting issues and correctly capitalized throughout? \quad \textbf{2.} Are all sentences grammatically correct and free from errors? \quad \textbf{3.} Are verb tenses used consistently within each section? \quad \textbf{4.} Do pronouns agree in number and gender with their antecedents? \quad \textbf{5.} Does the report avoid unnecessary repetition across sections when that repetition does not add comparison, evidence, or synthesis? \quad \textbf{6.} Does the report avoid elaborations that are tangential to the paper or its novelty evaluation? \quad \textbf{7.} When technical details recur across sections, are they used for a new analytical purpose rather than copied verbatim? \quad \textbf{8.} Is the language precise enough for academic evaluation while preserving necessary technical and comparative detail? \quad \textbf{9.} Is the report easy to read, without unnecessary complexity? \quad \textbf{10.} Are sentences generally concise, avoiding unnecessary wordiness? \quad \textbf{11.} Is terminology used consistently to avoid reader confusion? \quad \textbf{12.} Are paragraphs organized in a way that makes the argument easy to follow? \quad \textbf{13.} Does the report maintain a clear academic tone throughout? \quad \textbf{14.} Are section and subsection labels clear enough for readers to locate the relevant analysis quickly?
\end{tcolorbox}
\end{minipage}
\caption{Fluency checklist questions used in checklist evaluation.}
\label{fig:checklist-fluency}
\end{figure*}

\begin{figure*}[p]
\centering
\begin{minipage}{0.97\textwidth}
\begin{tcolorbox}[enhanced,colback=gray!10,colframe=black,coltitle=white,fonttitle=\bfseries,title={Evaluation Checklist: Effectiveness (17 questions)},boxrule=1pt,arc=2pt,left=6pt,right=6pt,top=6pt,bottom=6pt,fontupper=\footnotesize]
\setlength{\parindent}{0pt}\setlength{\parskip}{0pt}
\textbf{1.} Is the current report strictly divided into sections as required: 1. Paper Content Summary, 2. Point-wise Novelty Analysis, 3. Novelty Summary \quad \textbf{2.} Are there no extra sections or headers beyond the three required sections? \quad \textbf{3.} Does the 'Novelty Summary' (Section 3) draw its conclusions logically from the evidence and analysis presented in the preceding sections? \quad \textbf{4.} Does Section 3 clearly connect its final judgment to the point-wise analyses in Section 2? \quad \textbf{5.} In the 'Point-wise Novelty Analysis' (Section 2), does the analysis in subsection 2.d ('Details of Unique Differences') logically follow from and focus exclusively on the unique aspects identified in its corresponding subsection 2.c? \quad \textbf{6.} Does the analysis within subsection 2.d ('Details of Unique Differences') directly address the required questions for the specific innovation 'Classification' defined in subsection 2.a? \quad \textbf{7.} Does Section 2 maintain a clear organization between identified novelty points and their corresponding comparative analyses? \quad \textbf{8.} Within Section 2, does the report correctly use the required subsection headers: 'a. Claimed Novelty', 'b. Similarities', 'c. Unique Differences', and 'd. Details of Unique Differences' (if applicable)? \quad \textbf{9.} Does each sub-section under each section contain the required content in the template? \quad \textbf{10.} Does Section 3 end with a single-line summary of the innovation impact? \quad \textbf{11.} In Section 3, was the article's originality rated on a scale of 1 to 4 as required by the template? \quad \textbf{12.} Does the report clearly distinguish genuine contributions from less original or incremental aspects when making the final novelty judgment? \quad \textbf{13.} Does the final one-line summary include both an innovation score/rating and a concise explanation of the main strength and limitation? \quad \textbf{14.} Does each piece of content strictly adhere to the proposed theme without any signs of deviation? \quad \textbf{15.} When broader domain history or related-work background is included, is it directly relevant to evaluating the paper's novelty rather than tangential? \quad \textbf{16.} Does the entire report remain focused on analyzing the specified paper without introducing irrelevant topics or tangential discussions? \quad \textbf{17.} Does every section of the report address its designated purpose (content summary, point-wise novelty analysis, or novelty evaluation) without mixing or conflating objectives?
\end{tcolorbox}
\end{minipage}
\caption{Effectiveness checklist questions used in checklist evaluation.}
\label{fig:checklist-effectiveness}
\end{figure*}

\begin{figure*}[p]
\centering
\begin{minipage}{0.97\textwidth}
\begin{tcolorbox}[enhanced,colback=gray!10,colframe=black,coltitle=white,fonttitle=\bfseries,title={Evaluation Checklist: Completeness (31 questions)},boxrule=1pt,arc=2pt,left=6pt,right=6pt,top=6pt,bottom=6pt,fontupper=\footnotesize]
\setlength{\parindent}{0pt}\setlength{\parskip}{0pt}
\textbf{1.} Does the Paper Content Summary include the main objectives, methodology, and abbreviations that are mentioned in the provided text from the Main paper? \quad \textbf{2.} Does the summary explicitly state the paper's main objective in a clear, single sentence, if this objective is mentioned in the provided text? \quad \textbf{3.} Does the summary restate the specific problems the Main paper claims to solve, as described in the provided text? \quad \textbf{4.} Does the summary enumerate the key methodological components that are described in the provided text from the Main paper? \quad \textbf{5.} Does the summary define the uncommon abbreviations that appear in the provided text from the Main paper? \quad \textbf{6.} Does the summary include the main paper's most important technical components without drifting into unrelated background? \quad \textbf{7.} In the Point-wise Novelty Analysis, does the report identify the main claimed novelty points explicitly mentioned in the retrieved text of the main paper? \quad \textbf{8.} Does the Point-wise Novelty Analysis section avoid overlooking central novelty points from the retrieved text or combining distinct major contributions in a way that obscures the evaluation? \quad \textbf{9.} Does the report include the main theoretical or conceptual contributions claimed by the Main paper when they are available? \quad \textbf{10.} For theoretical novelty points, does the report cover the central theorem, proof target, formal claim, or conceptual framework described by the Main paper? \quad \textbf{11.} In the similarity analysis section, does the report correctly identify similarities to the innovation point that are mentioned within the provided text from related papers? \quad \textbf{12.} If the provided text from related papers describes prior methods with the same core algorithmic idea, does the report identify them as similar? \quad \textbf{13.} If the provided text from related papers describes prior works sharing the same problem formulation or task, does the report identify them as similar? \quad \textbf{14.} If the provided text from related papers describes alternative approaches to the same limitation or gap, does the report identify them as similar? \quad \textbf{15.} Does the report cover similarities in datasets, benchmarks, tasks, or evaluation settings when the related-paper evidence supports them? \quad \textbf{16.} Does the report cover similarities in empirical observations, trends, or diagnostic findings when the related-paper evidence supports them? \quad \textbf{17.} Does the report cover related works that challenge the paper's claimed novelty, not only related works that support it? \quad \textbf{18.} Does the report cover related works that are similar in theoretical framing or assumptions when such evidence is available? \quad \textbf{19.} Does the report cover related works that are similar in empirical finding or diagnostic conclusion when such evidence is available? \quad \textbf{20.} In the Unique Differences section, are the claims of uniqueness free from contradictions when checked against the provided text from related papers? \quad \textbf{21.} Does the report avoid claiming a feature as a Unique Difference if the provided related-paper evidence describes a similar feature, even if different terminology is used? \quad \textbf{22.} For each major unique point claimed in the report, does the report cover whether similar items are found within the provided text from related papers? \quad \textbf{23.} After comparing with the provided text from related papers, does the report cover whether the uniqueness claims made in the report still appear to hold? \quad \textbf{24.} Does the report cover whether the assertion of uniqueness for a specific feature is supported by the absence of that feature in the provided texts from related papers? \quad \textbf{25.} Does the report cover differences in dataset scale, annotation, coverage, benchmark design, or evaluation method when those differences are central to novelty? \quad \textbf{26.} Does the report cover cases where a dataset, benchmark, or system is valuable but follows an established design pattern? \quad \textbf{27.} Does the report make clear which related papers support major unique-difference claims? \quad \textbf{28.} Does the report cover standard engineering or evaluation practices when the paper presents them as novelty? \quad \textbf{29.} Does the report avoid omitting a key limitation of prior work that the Main paper explicitly addresses? \quad \textbf{30.} Does the report cover differences in theoretical assumptions, proof targets, or analytical scope when those differences are central to novelty? \quad \textbf{31.} When a claimed difference is mainly a minor implementation change, simple combination, or standard extension, does the report explicitly reflect that limited originality?
\end{tcolorbox}
\end{minipage}
\caption{Completeness checklist questions used in checklist evaluation.}
\label{fig:checklist-completeness}
\end{figure*}

\begin{figure*}[p]
\centering
\begin{minipage}{0.97\textwidth}
\begin{tcolorbox}[enhanced,colback=gray!10,colframe=black,coltitle=white,fonttitle=\bfseries,title={Evaluation Checklist: Faithfulness (29 questions)},boxrule=1pt,arc=2pt,left=6pt,right=6pt,top=6pt,bottom=6pt,fontupper=\footnotesize]
\setlength{\parindent}{0pt}\setlength{\parskip}{0pt}
\textbf{1.} Does the 'Paper Content Summary' accurately reflect the paper's main task, objective, and methods, with all factual details being directly verifiable from the provided text of the Main paper? \quad \textbf{2.} Does the 'Paper Content Summary' contain no information that is not explicitly supported by the provided text from the main paper? \quad \textbf{3.} For each 'Claimed Innovation' subsection, do the stated classification, core idea, and key technical elements accurately represent the contributions as described in the provided text from the Main paper? \quad \textbf{4.} Are all factual details within the 'Claimed Innovation' subsections directly verifiable from the provided text of the main paper? \quad \textbf{5.} Does the report avoid describing a result as universal or general if the provided Main-paper evidence only supports a limited setting? \quad \textbf{6.} Are all 'Similarities' (b) and 'Unique Differences' (c) described in the 'Comparative Analysis' section exclusively based on and directly supported by the content found within the provided text from related papers? \quad \textbf{7.} Are all specific examples of similar objectives or mechanisms mentioned in the 'Similarities' subsections (b) precisely aligned with the information present in the provided text from related papers? \quad \textbf{8.} Are all specific contrasts elaborated in the 'Unique Differences' subsections (c) directly and accurately supported by the provided text from related papers? \quad \textbf{9.} Are the explanations and details provided for each 'Similarity' (b) and 'Unique Difference' (c) fully and accurately supported by the provided text from related papers? \quad \textbf{10.} When the report says two works share a dataset, task, benchmark, or evaluation setting, is that shared setting supported by the provided related-paper evidence? \quad \textbf{11.} When the report states that a related work reports a quantitative result, is the number or trend consistent with the provided evidence? \quad \textbf{12.} Does the report avoid treating retrieved snippets as saying more than they actually say when drawing similarity or difference conclusions? \quad \textbf{13.} Does the report avoid claiming a related work includes an ablation, experiment, proof, or benchmark that is not present in the provided evidence? \quad \textbf{14.} Does the report avoid using references only as decoration when the surrounding sentence makes a substantive factual claim? \quad \textbf{15.} When the report summarizes a cluster of related papers, does it avoid making the cluster sound more homogeneous than the evidence supports? \quad \textbf{16.} When the report discusses prior limitations, does it tie those limitations to the correct related work or group of related works? \quad \textbf{17.} Does the report avoid conflating multiple related papers into one source when attributing a method, result, or limitation? \quad \textbf{18.} Does the report connect cited related works to the specific innovation point being discussed rather than using them as generic background? \quad \textbf{19.} When the report discusses prior advantages, does it tie those advantages to the correct related work or group of related works? \quad \textbf{20.} Does the report correctly distinguish between the Main paper's own contributions and the existing baselines or prior work described in the text, avoiding misattribution of features? \quad \textbf{21.} Does the report accurately interpret the logical flow and causal relationships of the proposed method of the Main paper, ensuring that the mechanism is explained exactly as the authors intended without logic errors? \quad \textbf{22.} Are the conclusions and claims in the report faithful to the nuances of the Main paper, avoiding exaggeration of results or misinterpretation of the limitations stated in the text? \quad \textbf{23.} Are all 'Similarities' and 'Unique Differences' presented in the 'Comparative Analysis' section strictly derived from and explicitly supported by the content found within the provided text of related papers? \quad \textbf{24.} Is the 'Details of Unique Differences' for each innovation point solely based on the unique aspects and details explicitly present in the provided text, without additional interpretation? \quad \textbf{25.} Are any performance advantages or comparative benefits claimed within the report directly and explicitly supported by comparative evidence present in the provided text of related papers? \quad \textbf{26.} Does the report avoid saying that prior work lacks a component when the provided evidence indicates that it has a comparable component? \quad \textbf{27.} Does the report distinguish between a related paper's claimed contribution and the current report's evaluation of that contribution? \quad \textbf{28.} When the report later makes an overall judgment about a contribution, is that judgment consistent with the evidence and conclusions already presented in the Unique Differences subsection? \quad \textbf{29.} Are the interpretations in 'Details of Unique Differences' grounded in the identified unique aspects and connected to the available evidence, rather than introducing unrelated or unsupported claims?
\end{tcolorbox}
\end{minipage}
\caption{Faithfulness checklist questions used in checklist evaluation.}
\label{fig:checklist-faithfulness}
\end{figure*}

\begin{figure*}[p]
\centering
\begin{minipage}{0.97\textwidth}
\begin{tcolorbox}[enhanced,colback=gray!10,colframe=black,coltitle=white,fonttitle=\bfseries,title={Evaluation Checklist: Depth (27 questions)},boxrule=1pt,arc=2pt,left=6pt,right=6pt,top=6pt,bottom=6pt,fontupper=\footnotesize]
\setlength{\parindent}{0pt}\setlength{\parskip}{0pt}
\textbf{1.} Does the 'Paper Content Summary' include specific technical terminologies (e.g., specific model architectures, loss functions, mathematical formulations) rather than just high-level functional descriptions? \quad \textbf{2.} Does the report explicitly define the specific input data types, datasets, or benchmark environments used in the main paper? \quad \textbf{3.} When discussing empirical or application-oriented contributions, does the report specify the datasets, benchmarks, domains, or experimental settings involved rather than describing evaluation only generically? \quad \textbf{4.} For dataset or benchmark contributions, does the report describe the resource's scale, composition, annotation, task coverage, or evaluation purpose when available? \quad \textbf{5.} For empirical or analytical contributions, does the report identify the non-obvious finding rather than merely stating that experiments were conducted? \quad \textbf{6.} Does the report connect claimed theoretical contributions to formal objects such as assumptions, propositions, theorems, proofs, bounds, or derivations? \quad \textbf{7.} Does the report connect claimed dataset or benchmark contributions to concrete properties such as scale, diversity, annotation process, coverage, or evaluation method? \quad \textbf{8.} Does the report mention important objective functions, losses, constraints, assumptions, equations, or proof targets when they are central to the paper? \quad \textbf{9.} Does the report identify the datasets, benchmarks, domains, or experimental environments used by the main paper rather than describing evaluation generically? \quad \textbf{10.} In the 'Similarities' section, does the report describe specific shared mechanisms, equations, or structural elements between the main paper and the retrieved works, rather than just generic thematic overlaps? \quad \textbf{11.} If specific baselines, prior models, datasets, systems, or theoretical frameworks are mentioned, does the report compare the current paper against the strongest or closest ones rather than only convenient examples? \quad \textbf{12.} Does the comparative analysis distinguish substantive innovation from changes in terminology, scale, benchmark choice, implementation detail, or packaging? \quad \textbf{13.} When a cited related paper is described as different, does the report identify the exact missing, changed, or weaker component in that prior work that supports the contrast? \quad \textbf{14.} When multiple related papers are cited for one novelty point, does the report distinguish the role of each cited paper in the comparison rather than treating them as interchangeable background? \quad \textbf{15.} For the closest prior works, does the report discuss them in enough technical detail to justify whether the current contribution is substantially different or mainly incremental? \quad \textbf{16.} When making an originality judgment, does the report connect that judgment to concrete overlaps or extensions relative to cited prior work rather than relying on broad field-level statements? \quad \textbf{17.} When a related paper is cited for a dataset or benchmark, does the report identify the relevant dataset property, task, scale, or evaluation setting? \quad \textbf{18.} When a related paper is cited for a theoretical result, does the report identify the relevant theorem, assumption, framework, or analytical conclusion? \quad \textbf{19.} When a related paper is cited for an empirical finding, does the report identify the relevant observation, trend, metric, or experimental setup? \quad \textbf{20.} In the Similarities section, does the report identify shared system components, pipeline stages, or infrastructure choices when relevant? \quad \textbf{21.} Does the report make clear which cited paper supports each major difference claim? \quad \textbf{22.} Does the report explain why the similarity matters for evaluating novelty rather than merely noting that two works are related? \quad \textbf{23.} When the provided text highlights similarities between the main paper and prior work, does the report's final innovation assessment reflect these similarities, rather than simply repeating the author's claims of novelty? \quad \textbf{24.} Does the report evaluate innovation along multiple dimensions--such as implementation, similarity to prior work, and technical contribution--based on the evidence in the provided texts? \quad \textbf{25.} Does the report appear to autonomously assess the author's claims on the basis of evidence from the provided texts, rather than just repeating the author's words? \quad \textbf{26.} Does the innovative summary acknowledge at least one limitation, boundary condition, or negative aspect of the claimed innovations, if such information is inferable from the provided texts? \quad \textbf{27.} Does the report identify standard practice or routine engineering choices when the comparison evidence supports that judgment?
\end{tcolorbox}
\end{minipage}
\caption{Depth checklist questions used in checklist evaluation.}
\label{fig:checklist-depth}
\end{figure*}

\clearpage

\begin{figure*}[p]
\centering
\begin{minipage}{0.97\textwidth}
\begin{tcolorbox}[
    enhanced,
    colback=gray!10,
    colframe=black,
    coltitle=white,
    fonttitle=\bfseries,
    title={Checklist Evaluation Answering Prompt},
    boxrule=1pt,
    arc=2pt,
    left=6pt,
    right=6pt,
    top=6pt,
    bottom=6pt,
    fontupper=\footnotesize
]

\textbf{Task overview.}
The evaluator is instructed to read a provided report and a database containing information on related papers, and then answer \texttt{yes} or \texttt{no} to specific checklist questions. The questions correspond to a particular evaluation dimension of the report.

\textbf{Dimension specification.}
The prompt provides the target dimension in the form:
\texttt{\{dimension\} -- \{definition\}}, followed by dimension-specific conditions, denoted as \texttt{\{conditions\}}.

\textbf{Instructions.}
\begin{itemize}[leftmargin=*]
    \item Read the instructions thoroughly.
    \item Carefully read both the report and the database.
    \item Understand the checklist questions and the definition of the target dimension.
    \item Respond to each question with \texttt{yes} or \texttt{no}.
    \item Base each answer on a clear rationale.
    \item Follow the specified answer format.
\end{itemize}

\textbf{Evaluation focus note.}
The prompt clarifies that the report may contain different kinds of content, including statements that summarize the main paper and statements that compare it with retrieved or cited related works. The exact naming of a group is less important than its evaluation focus. The evaluator should use the provided \texttt{\{focus\_note\}} for the current question group.

\textbf{Equivalent naming note.}
When evaluating structure or section-specific questions, the evaluator should judge flexibly according to the actual function and meaning of a section or subsection. If two titles express nearly the same function but use different wording, they should be treated as equivalent rather than marked as missing merely because of wording differences. In particular:
\begin{itemize}[leftmargin=*]
    \item \textit{Comparative Analysis}, \textit{Point-wise Novelty Analysis}, and \textit{Point-wise Comparative Analysis} are equivalent titles for the second main section.
    \item \textit{Innovation Point} and \textit{Novelty Point} are equivalent titles for each point-level subsection.
    \item \textit{Claimed Innovation} and \textit{Claimed Novelty} are equivalent subsection titles.
    \item \textit{Details of Unique Difference}, \textit{Details of Unique Differences}, and \textit{Unique Feature Description} are equivalent subsection titles.
    \item \textit{Innovation Summary} and \textit{Novelty Summary} are equivalent titles for the final main summary section.
    \item \textit{Genuine Innovations} and \textit{Genuine Novelties} are equivalent subsection titles.
    \item \textit{Aspects Lacking Originality}, \textit{Aspects Lacking Substantial Originality}, and \textit{Aspects lacking substantial originality} are equivalent subsection titles.
    \item \textit{References} and markdown reference-link definitions are acceptable citation-support sections and should not be treated as extra analytical sections.
\end{itemize}

\textbf{Answer format.}
The evaluator must return one line per checklist question using the required format:
\texttt{Q1: [Your Answer]},
\texttt{Q2: [Your Answer]},
and so on, where each answer is \texttt{yes} or \texttt{no}.

\end{tcolorbox}
\end{minipage}
\caption{Core prompt used by the evaluator to answer checklist questions.}
\end{figure*}
\clearpage

\begin{figure*}[p]
\centering
\begin{minipage}{0.97\textwidth}
\begin{tcolorbox}[
    enhanced,
    colback=gray!10,
    colframe=black,
    coltitle=white,
    fonttitle=\bfseries,
    title={Point-wise Novelty Analysis Prompt},
    boxrule=1pt,
    arc=2pt,
    left=6pt,
    right=6pt,
    top=6pt,
    bottom=6pt,
    fontupper=\footnotesize
]

Please perform a detailed point-wise novelty analysis for the following novelty point extracted from the paper
\texttt{\{paper\_name\}}.
\texttt{Novelty Point \{point\_num\}: \{innovation\_point\}}
\textbf{INSTRUCTIONS:}
\begin{itemize}[leftmargin=*,nosep]
    \item Use the Novelty Point description to understand the current claim's content, methodology, evidence, and claimed contribution.
    \item Use the retrieved prior-work evidence to compare against concrete prior papers and to support cited claims.
    \item Use the historical prior-work context from the system message as historical comparison evidence. Do not name or describe this source in your output, but when you rely on a concrete historical paper's innovation content or evaluation, cite its provided Historical citation marker.
    \item Use the selected leaf-theme histories for detailed paper-level comparison, and use the broader parent-theme overall summary only to understand the larger development trajectory of the field.
    \item In Similarities and Unique Differences, do not only say that the current point is in the same broad theme. Explain the content-level overlap or difference: what mechanism, dataset, theoretical claim, empirical finding, task formulation, or system capability is shared or different.
    \item If the historical context suggests that the current point is largely a small variation of prior historical innovations, say this as a natural judgment about prior work, not as a statement about the context source.
    \item Keep the output structure stable. Output each of the four subsections exactly once. Do not repeat b/c/d sections. Do not add extra Section 2 headings inside your response.
    \item Do not include meta-commentary, correction notes, validation notes, or source-debugging notes.
    \item Forbidden output phrases include: ``expert knowledge'', ``expert memory'', ``memory system'', ``RAGFlow'', ``system message'', ``system context'', ``retrieved related texts'', ``provided materials'', ``corrected for'', and ``The original reference''.
\end{itemize}
\textbf{OUTPUT FORMAT:}
\textbf{a) Restatement:}\\
Rephrase the novelty point clearly and in detail based on the novelty description. Include methodology, architecture, dataset/benchmark, theoretical claim, empirical evidence, or system details when they are available.
\textbf{b) Similarities:}\\
Compare this novelty point with prior work. If no similarities are found, state: ``Based on the available prior-work evidence, no explicit similarities with existing work were identified.'' Otherwise, describe specific shared objectives, mechanisms, datasets, evidence types, or problem formulations.
\textbf{c) Unique Differences:}\\
Excluding the similarities in part (b), analyze what makes this novelty point different from prior work. Use specific details from the novelty point description. If no unique differences are found, state: ``Based on the available prior-work evidence, no unique differences were identified for this novelty point.''
\textbf{d) Details of Unique Differences:}\\
If genuinely unique aspects were identified in part (c), analyze their significance in one paragraph. Follow the innovation type: for methods, explain the structural/procedural difference and claimed benefit; for theory, explain the central claim and rigor; for systems, explain the solved problem and demonstrated practicality; for datasets/benchmarks, explain the resource gap and enabled capabilities; for empirical/analytical points, explain the central finding and experimental support; for tasks/applications, explain why the task/application is important and non-obvious. If part (c) found no uniqueness, leave only a brief sentence explaining that no distinct details can be added.
\textbf{Important Reminder:}
During Similarities, Unique Differences, and Details, compare the current novelty point against prior work. Do NOT compare the novelty point with itself. Do NOT fabricate implementation details, metrics, datasets, or claims not present in the available evidence.
Now generate the analysis.

\end{tcolorbox}
\end{minipage}
\caption{Prompt used for point-wise novelty analysis against prior work.}
\end{figure*}
\clearpage

\begin{figure*}[p]
\centering
\begin{minipage}{0.97\textwidth}
\begin{tcolorbox}[
    enhanced,
    colback=gray!10,
    colframe=black,
    coltitle=white,
    fonttitle=\bfseries,
    title={Prompt used to generate the novelty summary},
    boxrule=1pt,
    arc=2pt,
    left=6pt,
    right=6pt,
    top=6pt,
    bottom=6pt,
    fontupper=\footnotesize
]

You are an Expert Novelty Assessment Specialist with deep expertise in evaluating the novelty, significance, and impact of research contributions. Your role is to synthesize point-wise comparative analyses into a balanced final novelty summary.
You will receive two inputs:
\begin{enumerate}[leftmargin=*,nosep]
    \item Section 2: point-wise novelty analysis for each claimed innovation.
    \item Historical prior-work context corresponding to the innovation points.
\end{enumerate}
\textbf{How to use the historical context:}
\begin{itemize}[leftmargin=*,nosep]
    \item Use it to understand how similar historical innovations were evaluated, including their ratings, assessment text, and historical standing.
    \item Use these historical evaluations as reference examples for calibrating whether the current paper is transformative, significant, incremental, or weak.
    \item Distinguish between claimed innovation points that have little substantial originality and claimed innovation points that are genuinely innovative.
    \item \ldots
    \item Do not name this context source or expose internal pipeline mechanics.
\end{itemize}
\textbf{Important restrictions:}
\begin{itemize}[leftmargin=*,nosep]
    \item DO NOT use any citations or references in Section 3.
    \item Base your analysis only on Section 2 and the historical context below.
    \item Use general terms like ``existing work'', ``prior methods'', ``historical innovations'', and ``related approaches''.
    \item \ldots
\end{itemize}
\textbf{Section 2 Content:}

\texttt{\{draft\_section2\}}
\textbf{Historical Context:}

\texttt{\{expert\_knowledge\_bundle\}}
\textbf{Generate Section 3 exactly as follows:}

\texttt{\#\# 3. Novelty Summary}

Write a comprehensive, detailed, and faithful synthesis of the paper's novel contributions. First articulate the paper's overall innovative characteristics. Then distinguish genuinely innovative contributions from aspects that lack substantial originality.
\textbf{Required content:}
\begin{itemize}[leftmargin=*,nosep]
    \item \textbf{Genuine Innovations:} Identify the innovation point(s) that are genuinely innovative, explain the core unique idea, and discuss its significance in historical context.
    \item \textbf{Aspects Lacking Originality:} Identify claimed innovations that are not substantially original. Categorize each as Direct Repetition, Minor Adjustment, Simple Combination, Standard Practice, or Other.
\end{itemize}
\textbf{Final One-line Summary:}

End with exactly one final one-line summary in this format: \texttt{**<score> - <label>:** <one sentence explaining the overall novelty, strengths, and limitations>.}
The score should be calibrated using both Section 2 and the historical evaluation examples.
\textbf{Rating rules:}
\begin{itemize}[leftmargin=*,nosep]
    \item \textbf{4 - Excellent / Transformative:} Entirely new question, highly novel methodology, or surprising perspective.
    \item \textbf{3 - Good / Significant:} Clear and substantial progress through a non-obvious methodology or creative combination.
    \item \textbf{2 - Fair / Incremental:} Limited extension through minor improvements, tweaks, or known techniques.
    \item \textbf{1 - Poor / Insufficient:} Lacks clear original contributions or presents well-known, trivial, or previously published ideas.
\end{itemize}
\textbf{Strict output rules:}
\begin{itemize}[leftmargin=*,nosep]
    \item Output ONLY Section 3 content, starting with ``\#\# 3. Novelty Summary''.
    \item The final line must include the score, label, and one explanatory sentence.
    \item Be concise, neutral, and faithful to Section 2 and the historical context.
    \item DO NOT use citations or paper-reference markers.
    \item \ldots
\end{itemize}

\end{tcolorbox}
\end{minipage}
\caption{Prompt used to generate the novelty summary.}
\end{figure*}
\clearpage

\begin{figure*}[p]
\centering
\begin{minipage}{0.97\textwidth}
\begin{tcolorbox}[
    enhanced,
    colback=gray!10,
    colframe=black,
    coltitle=white,
    fonttitle=\bfseries,
    title={Memory Development-History Prompt},
    boxrule=1pt,
    arc=2pt,
    left=6pt,
    right=6pt,
    top=6pt,
    bottom=6pt,
    fontupper=\footnotesize
]

\textbf{System prompt.}
You are an expert in structuring the development history of academic-paper innovation points. You are skilled at integrating publication time, paper titles, summaries, innovation claims, final decisions, and peer-review signals into a precise, high-detail, structured expert record. Your output will later be used for visualization, analysis, and as expert knowledge for another LLM to judge the novelty of a new innovation point and compare it against prior work. Therefore, you must preserve concrete technical content, accurately describe what each innovation point actually is, make the evolution of the category clear over time, and output a simple JSON object. You must cover every innovation point and must not omit any point. You should use review information selectively and critically: prioritize PC/AC meta-review over conflicting reviewer comments, use reviewer scores together with comment language and final decision to judge historical standing, ignore review comments that are irrelevant to the specific innovation point being discussed, and reason independently when the review evidence is absent, weak, or clearly unrelated.
\textbf{User prompt.}
You are given one category of academic innovation points, together with their source documents, publication years, paper titles, paper summaries, final decisions, and review comments.
\textbf{Category metadata:}
\begin{itemize}[leftmargin=*,nosep]
    \item Theme ID: \texttt{\{theme\_id\}}
    \item Theme Title: \texttt{\{theme\_title\}}
    \item Theme Description: \texttt{\{theme\_description\}}
\end{itemize}
\textbf{Your task:}
\begin{enumerate}[leftmargin=*,nosep]
    \item Output ONLY a JSON object. Do not output markdown, explanation, or prose outside JSON.
    \item The JSON object must follow this schema exactly:
\end{enumerate}

\textbf{Required JSON schema:} \texttt{\{theme\_overview; innovation\_points: [\{item\_number, year, paper\_title, innovation\_content, evaluation: \{rating, assessment\}\}]; overall\_evolution\_summary\}.}

\begin{enumerate}[leftmargin=*,resume,nosep]
    \item The \texttt{innovation\_points} list must be in chronological order as much as possible.
    \item For EACH innovation point, explain its concrete content in enough detail for downstream comparison, following the original innovation-point text closely, and assess its historical standing in the line of work.
    \item Use the paper summary, paper title, and innovation-point text as the primary evidence; use review comments, reviewer scores, meta-review, and final decision as auxiliary evidence for historical standing and significance, not as blind substitutes for reasoning.
    \item \ldots
\end{enumerate}
\textbf{Important constraints:}
\begin{itemize}[leftmargin=*,nosep]
    \item Only use the provided source material.
    \item Preserve chronological order as much as possible.
    \item The result should be structured JSON, not prose outside JSON.
    \item Do not include extra keys beyond the schema above.
    \item The history should be detailed enough to support downstream novelty judgment, side-by-side comparison, timeline visualization, and expert-memory use.
    \item Do not say that some points were omitted, merged, or unavailable.
    \item When using review signals, do not quote them mechanically without interpretation; integrate them into your own expert judgment.
\end{itemize}
\textbf{Source material:}

\texttt{\{source\_material\_text\}}

\end{tcolorbox}
\end{minipage}
\caption{Prompt used to write leaf-level development histories in the historical research memory.}
\end{figure*}
\clearpage

\end{document}